\documentclass[10pt,twocolumn,letterpaper]{article}
\ifdefined\pdfsuppressptexinfo\pdfsuppressptexinfo=-1\fi

\usepackage[pagenumbers]{cvpr} %

\usepackage{cuted}
\usepackage{needspace}
\usepackage{currfile}
\usepackage{caption}
\usepackage{graphicx}
\usepackage{amsmath,amssymb}
\usepackage{booktabs}
\usepackage{multirow}
\usepackage{xcolor}
\usepackage{pgfplots}
\pgfplotsset{compat=1.18}
\usetikzlibrary{calc}
\usepgfplotslibrary{groupplots}

\usepackage{colortbl}
\DeclareMathOperator*{\argmax}{arg\,max}

\newcommand{\plus}{$+$\hphantom{0}}

\definecolor{cRead}{HTML}{D65F7D}    %
\definecolor{cReadB}{HTML}{7795B8}   %
\definecolor{cDirect}{HTML}{485992}  %
\definecolor{cSelf}{HTML}{A0A8C0}    %
\definecolor{cGray}{HTML}{7F7F7F}

\definecolor{ours}{RGB}{232,240,254}

\pgfplotsset{
  odplot/.style={
    tick label style={font=\scriptsize},
    label style={font=\footnotesize},
    legend style={font=\scriptsize, draw=none, fill=none},
    legend cell align=left,
    grid=major, grid style={gray!18},
    axis line style={gray!60},
    tick style={gray!60},
  }
}

\definecolor{cvprblue}{rgb}{0.21,0.49,0.74}
\usepackage[pagebackref,breaklinks,colorlinks,allcolors=cvprblue]{hyperref}

\hypersetup{pdftitle={Hindsight Unlocks Robot Test-Time Compute},pdfauthor={Sunwoo Hong, Ali Pesaranghader, Se-Young Yun},pdfsubject={Hindsight-supervised action verification}}

\def\confName{CVPR}
\def\confYear{2027}

\title{Hindsight Unlocks Robot Test-Time Compute}

\author{
  Sunwoo Hong$^1$, \quad Ali Pesaranghader$^{2,\dagger}$, \quad Se-Young Yun$^{1,\dagger}$ \\
  $^1$KAIST AI \qquad $^2$LG Electronics, Toronto AI Lab \\
  {\tt\small sunwoo@kaist.ac.kr} \\
  {\small\href{https://sunwoohong.github.io/r2v/}{\color{cDirect}\textbf{\texttt{https://sunwoohong.github.io/r2v/}}}}
}

\begin{document}
\raggedbottom
\maketitle
\let\thefootnote\relax\footnotetext{$^\dagger$Corresponding authors.}
\begin{strip}
\centering
\vspace{-0.8em}
\includegraphics[width=\linewidth]{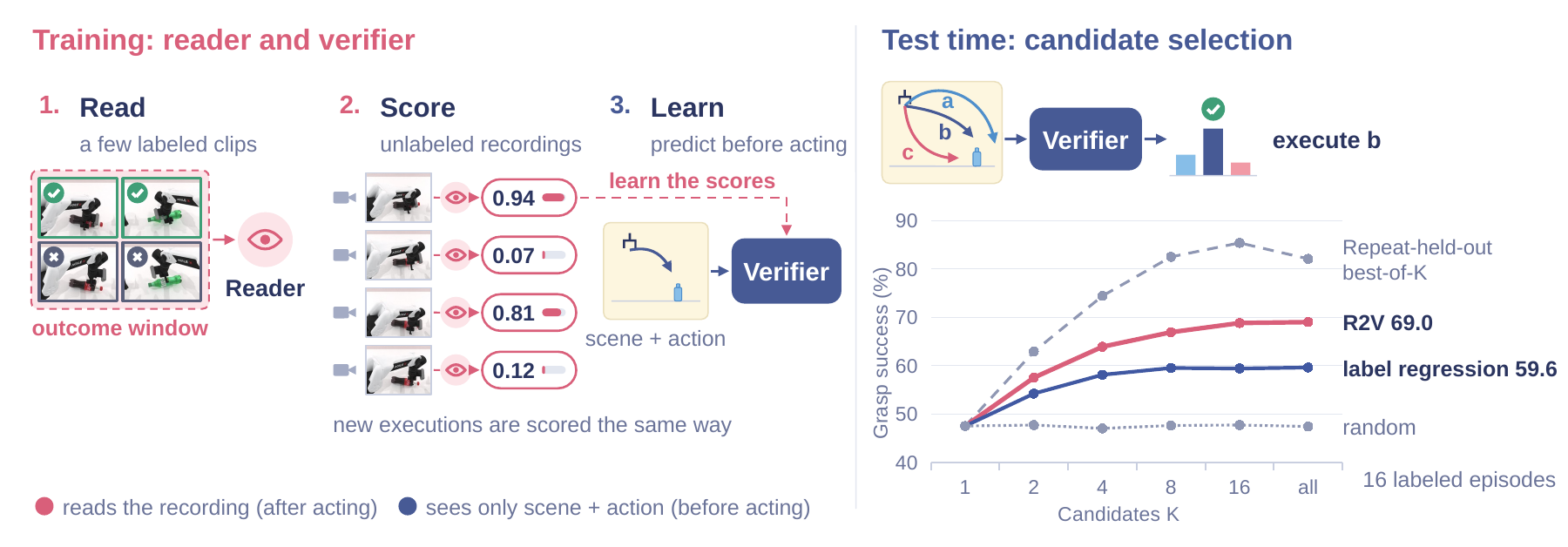}
\captionsetup{hypcap=false}
\captionof{figure}{\textbf{R2V training and action selection.} We fit a reader on labeled clips and use its scores on further executions to train a verifier with scene and action inputs. The verifier ranks candidate actions before execution. Right: tree-verifier success versus candidate count on the 195-scene evaluation set of \cref{fig:tts}, using 16 labeled episodes.}
\captionsetup{hypcap=true}
\label{fig:teaser}
\end{strip}

\begin{abstract}
Robot planners and generative policies can propose many candidate actions, but the robot must decide before execution which one will succeed. An action's outcome is far easier to recognize from its execution recording than to predict from the scene and plan. On dynamic grasping, a reader of a short outcome clip reaches AUC 0.85 from 16 labeled source episodes, while a pre-action predictor reaches 0.70 using all 158. Generated futures do not close this gap. In a pushing diagnostic, a readout that identifies the best action perfectly from simulator-rendered futures performs near chance on DreamDojo generations. Read-to-Verify (R2V) therefore uses recorded executions to train a pre-action verifier. A reader fitted on labeled executions scores additional recordings, and a verifier predicts those scores from the scene and proposed action alone. With the action generator fixed, a tree-based R2V verifier raises grasp success from 47.5\% with one candidate to 69.0\% with the full pool, versus 59.6\% for direct supervision, and the improvement holds for both tree and neural verifiers on the full evaluation set. With 16 labeled source episodes and additional unlabeled recordings, a neural verifier matches direct supervision on 128 labeled episodes. Recording-window and self-training controls attribute the gain to outcome information in the recordings.
\end{abstract}
\section{Introduction}\label{sec:intro}

A robot grasping a moving bottle must choose an approach and decide when to close its gripper. A motion planner can generate several feasible trajectories, but the robot still has to predict which one will hold the bottle. We use recorded executions to learn this judgment and test whether it lets the robot benefit from sampling more candidate plans.

In our grasping experiment, a randomly chosen trajectory succeeds in 47.5\% of scenes. Selecting among 16 candidates using two execution repeats and evaluating on a third gives 85.4\%. This reference uses outcomes unavailable before acting, but it shows that the candidate pools contain substantially better choices than a random trajectory. A verifier trained directly on sparse success labels recovers only part of that difference, even as we give it more candidates to choose from.

The recording of an executed grasp offers a useful training signal. After the gripper closes, a short clip shows whether the bottle stays in the hand or falls away. A linear reader of this clip reaches AUC 0.85 from only 16 labeled source episodes (about 280 grasp executions), while our pre-action predictor reaches 0.70 with all 158 (\cref{fig:auc}). We call this difference the \emph{foresight gap}. The predictor must infer the result of contact from the scene and planned motion, while the reader observes what happened.

An alternative is to generate future observations and judge each candidate from those rollouts. In a pushing diagnostic, a readout that identifies the best action from simulator-rendered futures in all 48 scenes selects it only 27.1\% of the time from DreamDojo samples, close to the 25\% chance rate. Outcome readers likewise separate recorded rollouts more accurately than the imagined observations of Cosmos Policy and Fast-WAM, though at different horizons. This motivates using actual executions as training evidence and learning a selector that needs no future frames at test time.

\textbf{Read-to-Verify (R2V)} uses an outcome reader to supervise the pre-action verifier (\cref{fig:teaser}). We fit the reader on labeled recordings, apply it to the remaining executions, and regress its scores from the corresponding scene--action inputs. At test time, the verifier ranks proposed actions without seeing their future outcomes. Once the reader has been fitted, further recorded attempts provide training targets without additional success annotation.

The better verifier changes the value of additional sampling. With the trajectory generator and verifier weights fixed, the tree R2V verifier trained on 16 labeled episodes climbs from 47.5\% with one candidate to 69.0\% with the full pool, while direct regression reaches 59.6\%. Beyond four candidates, R2V gains another 5.1 points, compared with 1.5 for direct regression. For the neural verifier, direct regression peaks at two candidates, then falls to 43.7\%, below random selection, whereas R2V continues to improve. 

In summary:
\begin{itemize}
  \item We identify the \textbf{foresight gap} and introduce \textbf{R2V}, which uses outcome recognition to supervise a pre-action verifier from additional recorded executions.
  \item On 512 evaluation scenes, R2V raises grasp success from 69.9\% to \textbf{75.5\%} with tree verifiers and from 59.2\% to \textbf{69.4\%} with neural verifiers. With 16 labeled episodes, both R2V verifiers outperform direct supervision on 32.
  \item With \textbf{16 labeled episodes} and 1,080 unlabeled single-execution recordings, R2V matches direct supervision on \textbf{128 labeled episodes}. Recording-window and self-training controls trace the gain to outcome information in the recordings.
\end{itemize}

We further show that a pretrained vision-language model can provide the reader scores, and report results on four additional manipulation tasks. A preliminary closed-loop PushT extension is reported in \cref{app:closedloop}.
\section{Related Work}\label{sec:related}

\paragraph{Action selection.}
Levine et al.~\cite{levine2016handeye} learn grasp-success predictions from images and candidate motions, and QT-Opt learns a Q-function for long-horizon, closed-loop grasping~\cite{kalashnikov2018qtopt}. More recent selectors use offline reinforcement-learning values to guide generalist policies, as in V-GPS and Hume~\cite{pmlr-v270-nakamoto25a, song2025hume}. RoboMonkey trains a vision-language verifier on synthetic comparisons~\cite{kwok2025robomonkey}. Concurrent work trains a compact verifier from terminal outcomes of frozen-policy rollouts~\cite{hsieh2026verification}. In R2V, we obtain that supervision by reading completed executions. Our candidate-budget evaluation follows the sampling-and-selection setting used in language-model inference and sampling-based control~\cite{snell2024scaling,deboer2005cem,williams2017mppi}.

\paragraph{Training with recorded outcomes.}
Success detectors and video-based reward models derive success or reward signals from observations~\cite{du2023success,escontrela2023viper,ma2023vip,lee2026roboreward}. Outcome supervision and hindsight relabeling are also used to improve policies from experience~\cite{pi2025pistar,luo2024hilserl,xu2026lfh,andrychowicz2017her}. R2V has a teacher--student structure in which the teacher sees information absent from the student's test-time inputs, connecting it to privileged-information learning and distillation~\cite{vapnik2009lupi,lopezpaz2016unifying,hinton2015distilling,chen2019cheating}. PUMA, introduced with the DOMINO benchmark, uses world queries to forecast object-centric future states for dynamic manipulation~\cite{fang2026domino}, while Privileged Foresight Distillation transfers future-aware corrections to a world action model~\cite{fang2026pfd}. We use the recorded outcome to estimate a scalar target for action selection. Direct regression and self-training~\cite{xie2020noisy,krzywicki2023grasping} provide comparisons with the same pre-action inputs but different supervision.

\paragraph{Prediction for control.}
Visual foresight evaluates actions through predicted observations~\cite{finn2017deep,ebert2018visual}, while DINO-WM and V-JEPA~2-AC plan in learned feature spaces~\cite{zhou2025dinowm,assran2025vjepa2}. UniPi generates a video plan and extracts actions from it~\cite{du2023unipi}. World action models predict future video alongside actions~\cite{li2025uva,ye2026dreamzero,yuan2026fastwam}, though some skip it at inference~\cite{li2025uva,yuan2026fastwam}. Cosmos Policy ranks candidates with a learned value~\cite{kim2026cosmospolicy}, and GeoBoN ranks imagined rollouts by geometric consistency~\cite{zhao2026geobon}. WCD ranks candidates with internal generative signals and learns an online predictor from the mismatch between imagined and realized observations~\cite{zhang2026wcd}. R2V instead predicts task success from the scene and proposed action, supervised by outcomes read from recorded executions, without generating futures. Prediction accuracy and decision quality are distinct objectives in model-based control~\cite{lambert2020objective,grimm2020value, pmlr-v54-farahmand17a}. Our outcome-feature predictor estimates the representation used by the reader, providing an alternative to predicting its scalar score. We separately evaluate action ranking from generated video using DreamDojo~\cite{gao2026dreamdojo} and the world action models above.

\section{The Foresight Gap}\label{sec:asym}

We compare recognition and prediction on the same grasp outcomes. The reader sees a 1.2-second recording, whereas the predictor receives only the scene and planned action. Each learning curve uses the same labeled episodes at a given budget and is evaluated on 32 separate development scenes.

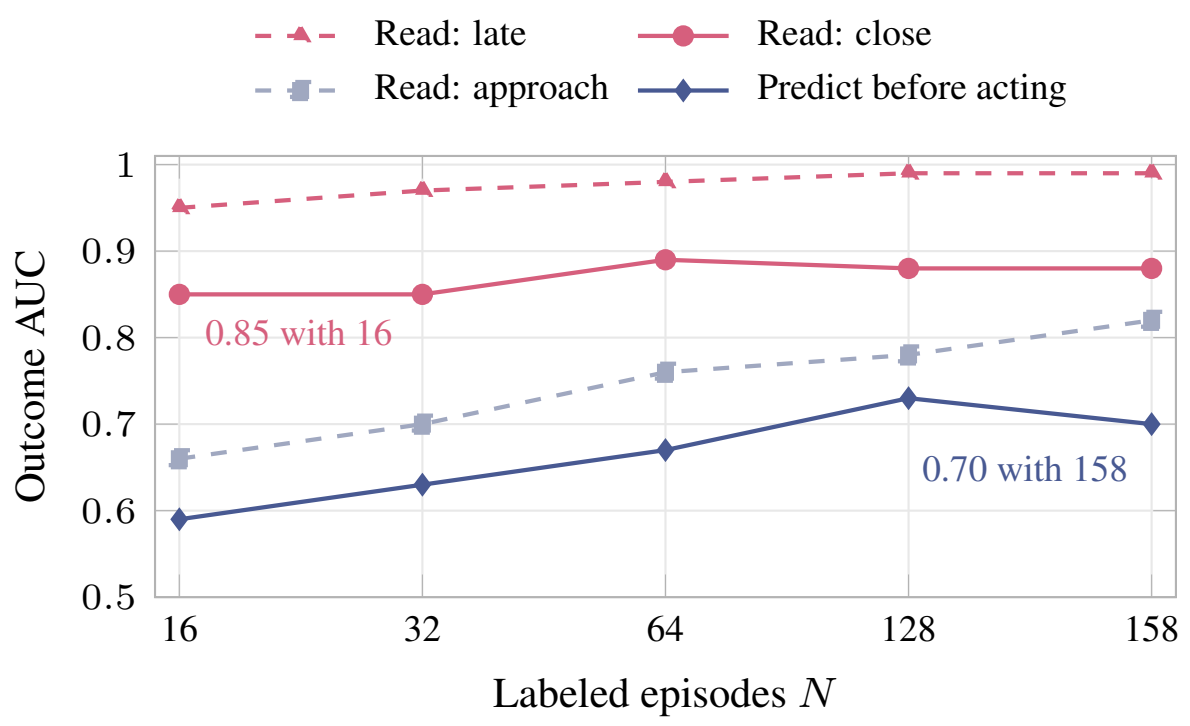
\begin{figure}[t]
\centering
\begin{tikzpicture}
\begin{axis}[
  odplot,width=\linewidth,height=4.5cm,
  xmin=-0.1,xmax=4.1,ymin=0.5,ymax=1.01,
  xtick={0,1,2,3,4},xticklabels={16,32,64,128,158},
  ytick={0.5,0.6,0.7,0.8,0.9,1.0},
  xlabel={Labeled episodes $N$},ylabel={Outcome AUC},
  legend columns=2,legend style={at={(0.5,1.06)},anchor=south,
  font=\scriptsize,draw=none,fill=none,column sep=3pt},
  tick label style={font=\scriptsize},label style={font=\footnotesize},
]
\addplot[cRead,dashed,mark=triangle*,thick,mark size=1.6pt] coordinates {(0,0.95) (1,0.97) (2,0.98) (3,0.99) (4,0.99)};
\addlegendentry{Read: late}
\addplot[cRead,mark=*,thick,mark size=1.6pt] coordinates {(0,0.85) (1,0.85) (2,0.89) (3,0.88) (4,0.88)};
\addlegendentry{Read: close}
\addplot[cSelf,dashed,mark=square*,thick,mark size=1.6pt] coordinates {(0,0.66) (1,0.7) (2,0.76) (3,0.78) (4,0.82)};
\addlegendentry{Read: approach}
\addplot[cDirect,mark=diamond*,thick,mark size=1.6pt] coordinates {(0,0.59) (1,0.63) (2,0.67) (3,0.73) (4,0.7)};
\addlegendentry{Predict before acting}
\node[font=\scriptsize,text=cRead,anchor=north west] at (axis cs:0.03,0.842) {0.85 with 16};
\node[font=\scriptsize,text=cDirect,anchor=north east] at (axis cs:3.98,0.685) {0.70 with 158};
\end{axis}
\end{tikzpicture}
\caption{\textbf{Reader and pre-action predictor AUC.} AUC for grasp success on 32 development episodes. The close-window reader reaches 0.85 with 16 labeled episodes. The pre-action predictor reaches 0.70 with 158. The recorded windows and label subsets are matched across the reader curves.}
\label{fig:auc}
\end{figure}

A linear probe on the gripper-closing clip reaches AUC 0.85 with 16 labeled episodes. The pre-action predictor reaches 0.59 at that budget and 0.70 with all 158 episodes (\cref{fig:auc}). Moving the reader's window 0.6 seconds later raises its AUC to 0.95 at $N=16$. By contrast, the approach window ends before gripper closure and gives AUC 0.66, despite showing the same execution. The difference between reading and prediction depends on which part of the interaction is visible.

Repeated executions help distinguish this learning difficulty from outcome variability. Two executions of a candidate predict a third with AUC 0.975, so much of the outcome is repeatable. The pre-action learner nevertheless has to infer that outcome from the scene and trajectory. R2V uses reader scores as targets for this prediction. In the held-out example in \cref{fig:qual}, direct regression ranks the candidate that fails in all three repeats above the one that succeeds in all three, while the R2V verifier gives the opposite ranking.

\section{Read-to-Verify}\label{sec:method}

\subsection{Reader and verifier training}
For scene $h$, let $\mathcal A(h)$ be the candidate actions and let $X=x(h,a)$ describe the scene and a proposed action $a$. Executing the action produces a recorded window $V$ and an outcome $Y\in\{0,1\}$. We observe outcomes for executions from $N$ labeled episodes, denoted by $\mathcal E_{\mathcal L}$. The remaining set $\mathcal E_{\mathcal U}$ contains scene--action inputs and recordings without outcome labels. At test time, the verifier selects $\hat a=\argmax_{a\in\mathcal A(h)}s_\theta(x(h,a))$.

\paragraph{Conditional outcome prediction.}
Write $\eta(X,V)=\mathbb E[Y\mid X,V]$. Conditioning on the pre-action input gives
\begin{equation}\label{eq:identity}
f(X)=\mathbb E[Y\mid X]=\mathbb E[\eta(X,V)\mid X].
\end{equation}
Thus the success probability before execution is the conditional expectation of an outcome estimate that also uses the recording. When $Y\perp X\mid V$, this estimate depends only on the video, $g^\star(V)=\mathbb E[Y\mid V]$. We approximate it with a reader fitted on labeled recordings and use the reader's outputs to train the verifier on the larger execution pool.

\paragraph{Reader.}
We encode each recorded window with a frozen video encoder $\phi$ and fit a logistic reader $g_\psi(\phi(V))$ using outcomes from $\mathcal E_{\mathcal L}$. The fitted reader scores recordings in both training sets. Its logit $q(h,a)$ is standardized using the mean and standard deviation of the verifier's training pool to obtain $\bar q(h,a)$.

\paragraph{Verifier.}
We train $s_\theta$ by regressing these targets from the corresponding scene--action inputs:
\begin{equation}\label{eq:verifier}
\min_\theta\sum_{(h,a)\in\mathcal E_{\mathcal L}\cup\mathcal E_{\mathcal U}}
\big(s_\theta(x(h,a))-\bar q(h,a)\big)^2.
\end{equation}
A target is paired only with the action that produced its recording. Inference requires the scene and candidate descriptions but neither the reader nor future frames. We execute the highest-scoring feasible action.

The population identity in \cref{eq:identity} uses probabilities. We regress standardized logits in the implementation to retain differences between confident readings, then use the predictions as ranking scores. \Cref{app:trees} compares hard, probability, and logit targets under matched learners.

\paragraph{Label requirements.}
Recognizing an outcome and predicting its dependence on the scene and action can require different numbers of labels, even when $V$ is determined by $X$. The parity construction in \cref{app:theory} gives an explicit example. At fixed noise and confidence, a reader needs a constant number of labels to identify a one-bit mapping from $V$ to $Y$, while direct prediction requires $\Omega(d)$ labels for a $d$-dimensional relation. Unlabeled pairs $(X,V)$ allow the prediction to be learned without annotating every outcome. The same analysis separates reader error that is predictable from $X$ from error that averages out in the conditional expectation.

\paragraph{Baselines.}
Direct \emph{label regression} uses the verifier architecture and squared loss in \cref{eq:verifier}, trained on the observed outcomes in $\mathcal E_{\mathcal L}$. For \emph{self-training}, we fit a pre-action predictor on that set, predict targets for $\mathcal E_{\mathcal U}$, and refit using both sets. Self-training therefore has access to the same unlabeled scene--action inputs as R2V but does not use their recordings.

\subsection{Candidate opportunity and selection loss}
For a pool of $K$ actions, let $a^\star$ maximize the true success probability $f$ and let $\hat a$ maximize the verifier's score. The gain over a uniformly chosen candidate is
\begin{equation}\label{eq:selection}
\mathbb E f(\hat a)-\mathbb E\tfrac1K\textstyle\sum_a f(a)
=\underbrace{H_K}_{\text{candidate opportunity}}
-\underbrace{\mathcal R_K}_{\text{selection loss}},
\end{equation}
where $H_K=\mathbb E[\max_a f(a)-\frac1K\sum_a f(a)]$ and
$\mathcal R_K=\mathbb E[f(a^\star)-f(\hat a)]$.
For a given pool, $H_K$ is the improvement available over random selection, and $\mathcal R_K$ is the amount lost by choosing $\hat a$ instead of $a^\star$. Our budget experiment evaluates each trained verifier at several $K$ values without updating its weights. A repeat-held-out best-of-$K$ selector estimates a reference performance from repeated executions. It selects using two repeats and is evaluated on the third. \Cref{app:theory} bounds selection loss in terms of score error and candidate count.

\subsection{Outcome features and single-execution logs}
We also train an \emph{outcome world model}, $m_\theta(X)\approx\mathbb E[\phi(V)\mid X]$, and score its predicted feature with the reader. For a linear reader logit $w^\top\phi(V)+b$,
\[
\mathbb E[w^\top\phi(V)+b\mid X]
=w^\top\mathbb E[\phi(V)\mid X]+b.
\]
Predicting the feature and then applying this reader has the same population score target as scalar regression. The finite-sample estimators can differ, however, because one predicts a feature vector and the other a single score. We compare them across label budgets.

Training does not require outcomes for every candidate in a scene. One executed action and its recording provide a pair $(X,V)$ for \cref{eq:verifier}. Our single-execution studies collect either a random action or the action chosen by a deployed selector in each new scene. We also replace the fitted reader with a pretrained vision-language model and use its output logit as the target.

\section{Experimental Setup}\label{sec:setup}

\paragraph{Dynamic grasping.}
We use the moving-bottle task in DOMINO~\cite{fang2026domino}. A bottle follows a sampled motion, and cuRobo~\cite{sundaralingam2023curobo} plans up to 32 grasp trajectories, with 18.6 feasible candidates per scene on average. The verifier input concatenates a frozen V-JEPA~2.1 ViT-B feature~\cite{murlabadia2026vjepa21} of the first second with a plan summary, arrival and closing times, and the commanded trajectory at 16 time points. We use the same encoder for the reader's 1.2-second clip.

\paragraph{Training data.}
The fit pool contains 158 source scenes and 2,797 executed candidates. We reveal outcomes for $N\in\{16,32,64,128,158\}$ source episodes and retain the other executions as unlabeled recordings. A labeled episode includes labels for its executed candidates, so $N=16$ corresponds to approximately 280 labeled grasps. Varying the revealed labels within a recorded pool separates the amount of outcome supervision from the amount of execution data. In the single-execution studies, each new scene contributes one recorded action.

We train gradient-boosted trees and a neural verifier, using identical inputs, labeled episode IDs, learner settings, and regression loss for R2V and direct regression. Results average five label subsets. The neural verifier ensembles three training seeds and selects epochs by cross-validation within the labeled training episodes. Reader and optimizer details are in \cref{app:details}.

\paragraph{Evaluation sets.}
The grasping evaluations use the following scene sets:
\par\smallskip
{\small
\setlength{\tabcolsep}{3pt}
\noindent\begin{tabular*}{\columnwidth}{@{\extracolsep{\fill}}lr@{}}
\toprule
Evaluation & Scenes \\
\midrule
Main selection (\cref{tab:main}) & 512 \\
Candidate-budget curves & 195 \\
Reader development & 32 \\
Window / budget screening & 128 \\
\bottomrule
\end{tabular*}}
\par\smallskip
The 195 budget scenes are a subset of the 512 confirmation scenes, with at least 16 feasible candidates, and are shared across all $K$. Each selected trajectory is scored by mean success over three stored executions. Methods selecting the same trajectory share those outcomes. Development and screening scenes are separate from confirmation scenes.

\paragraph{Uncertainty.}
We report paired differences and bootstrap intervals with explicit resampling units~\cite{NEURIPS2021_f514cec8}. Registered neural contrasts use a paired sign-flip test with Holm correction~\cite{holm1979}. Tree rate intervals resample five label subsets, while paired tree intervals resample scene-level differences. \Cref{app:protocol} specifies the evaluation and registered comparisons.

\paragraph{Other environments.}
We use place can basket, handover block, and stamp seal in RoboTwin~\cite{chen2025robotwin2}, and adjust bottle in DOMINO~\cite{fang2026domino}. Each scene has eight scripted contact or timing alternatives. Each task has 150 training and 100 screening scenes, with 400 confirmation scenes per RoboTwin task and 996 for adjust bottle. Neural results average 20 labeled subsets, and each evaluation candidate is executed three times. For the generated-future diagnostics, we use 48 PokeWorld~\cite{tan2026identifiability} scenes with four pushes and eight DreamDojo samples per push, along with rollouts from Cosmos Policy and Fast-WAM. \Cref{app:tasks,app:push} give the task-specific protocols.
\section{Action Selection Results}\label{sec:labels}

\begin{figure*}[t]
\centering
\includegraphics[width=\linewidth]{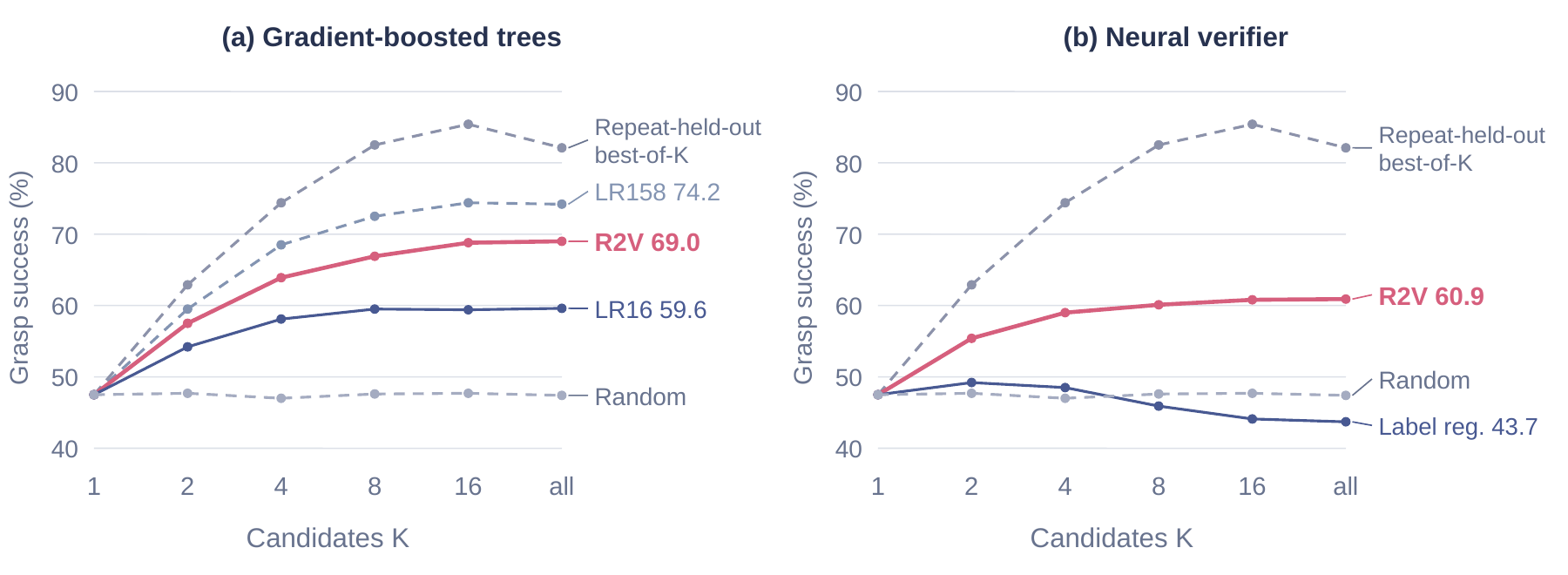}
\caption{\textbf{Grasp success as a function of candidate count.} We evaluate each learner on the same sampled pools on 195 confirmation scenes with at least 16 feasible candidates. Each point averages 50 candidate-set draws per scene and the labeled subsets. Repeat-held-out best-of-$K$ selects with two executions and is scored on a third. Finite-repeat selection need not yield a monotone reference curve. The full-label tree baseline is shown as a separate reference.}
\label{fig:tts}
\end{figure*}

\paragraph{Candidate budget.}
We evaluate the trained verifiers on candidate subsets of increasing size (\cref{fig:tts}). On the 195-scene budget set, tree R2V rises from 47.5\% at $K=1$ to 57.5\%, 63.9\%, 66.9\%, and 68.8\% at $K=2,4,8,16$, reaching 69.0\% with the full pool. Direct regression on the same 16 labeled episodes reaches 59.6\%. Beyond $K=4$, R2V gains 5.1 points, compared with 1.5 for direct regression. The repeat-held-out best-of-$K$ reference reaches 85.4\% at $K=16$.

The neural R2V verifier reaches 60.9\% with the full pool. Direct neural regression peaks at two candidates and falls to 43.7\%, below random selection. The tree baseline improves over random selection on the same scenes, so the decline does not occur for every directly supervised verifier.

\begin{table}[t]
\centering
\footnotesize
\setlength{\tabcolsep}{3pt}
\caption{\textbf{Grasp selection on 512 confirmation episodes.} Success (\%) averages three executions of the chosen candidate. Label regression (LR) and R2V share inputs and learner settings. The outcome world model uses an epoch choice fixed on 32 labeled development scenes. Intervals for the tree success rates resample five label subsets. Neural, VLM, and paired intervals resample episodes. VLM targets use a configuration selected on 32 labeled development episodes. The synchronization rule uses task geometry without fitting. The $\dagger$ contrasts were registered before execution and hold at Holm-adjusted $p<0.001$.}
\label{tab:main}
\begin{tabular}{@{}lcc@{}}
\toprule
Selector & $N=16$ & $N=32$ \\
\midrule
\multicolumn{3}{@{}l}{\emph{Gradient-boosted trees}} \\
Label regression & 69.9 [66.8, 72.8] & 73.5 [71.1, 76.2] \\
\rowcolor{ours} R2V & \textbf{75.5} [74.4, 76.5] & \textbf{77.6} [76.0, 79.0] \\
\midrule
\multicolumn{3}{@{}l}{\emph{Neural verifier}} \\
Label regression & 59.2 [55.9, 62.5] & 60.1 [57.0, 63.2] \\
Self-training & 57.1 [53.8, 60.5] & 58.6 [55.4, 61.8] \\
R2V, approach window & 61.0 [57.8, 64.1] & 62.1 [58.9, 65.2] \\
Outcome world model & 67.8 [64.6, 70.9] & 70.3 [67.0, 73.5] \\
\rowcolor{ours} R2V, close window & \textbf{69.4} [66.2, 72.5] & \textbf{71.6} [68.3, 74.7] \\
\midrule
\multicolumn{3}{@{}l}{\emph{Pretrained VLM reader}} \\
R2V, tree verifier & \multicolumn{2}{c}{76.4 [72.8, 79.8]} \\
R2V, neural verifier & \multicolumn{2}{c}{68.6 [64.7, 72.3]} \\
\midrule
\multicolumn{3}{@{}l}{\emph{Task-specific reference}} \\
Analytic synchronization & \multicolumn{2}{c}{71.0} \\
\midrule
\multicolumn{3}{@{}l}{\emph{Paired differences (percentage points)}} \\
Trees: R2V $-$ label regression & $+$5.6 [3.9, 7.4] & $+$4.1 [2.6, 5.5] \\
\multicolumn{3}{@{}l}{\emph{Neural verifier}} \\
R2V $-$ label regression$^\dagger$ & $+$10.2 [8.0, 12.3] & $+$11.5 [9.3, 13.7] \\
R2V $-$ self-training$^\dagger$ & $+$12.2 [9.9, 14.5] & $+$13.0 [10.6, 15.3] \\
Close $-$ approach$^\dagger$ & \plus8.4 [6.4, 10.4] & \plus9.5 [7.7, 11.3] \\
\bottomrule
\end{tabular}
\end{table}

\paragraph{Tree and neural verifiers.}
Across all 512 confirmation scenes, tree R2V improves success from 69.9\% to 75.5\%, a paired gain of 5.6 points [3.9, 7.4]. The neural improvement is 10.2 points [8.0, 12.3], from 59.2\% to 69.4\%. R2V with 16 labeled episodes exceeds direct regression with 32 for each learner (\cref{tab:main}). The task-specific synchronization rule achieves 71.0\%, below tree R2V and above neural R2V. Comparisons over 20 label subsets, a separate validation protocol, and DINOv2 or SigLIP features also retain gains over matched direct regression (\cref{app:budget,app:encoders}). Other direct-supervision losses and the synchronization score as an input feature are evaluated in \cref{app:baselines}.

\paragraph{Supervision ablations.}
Self-training reaches 57.1\% and 58.6\% at $N=16,32$, compared with 69.4\% and 71.6\% for R2V. It uses the same additional pre-action inputs but predicts their targets without video. In a second ablation, we retain the reader but train the verifier only on labeled executions. This labeled-only R2V model reaches 60.9\% on the screening set, compared with 61.6\% for direct regression, so the reader alone does not reproduce the improvement. Hard reader targets retain most of the gain in the matched target study (\cref{app:trees}).

\paragraph{Pretrained VLM reader.}
We ask Cosmos-Reason2 (8B) whether the bottle is held in the gripper in the late recording window and use its yes-token logit to supervise the verifier. The resulting tree and neural verifiers reach 76.4\% and 68.6\%. Verifier fitting uses these scores without success labels, although the VLM configuration is selected on 32 labeled development episodes (\cref{app:vlm}).

\paragraph{Outcome-feature prediction.}
The outcome world model predicts the reader's visual feature before applying the reader (\cref{sec:method}). It reaches 67.8\% and 70.3\% at $N=16,32$ (\cref{tab:main}), and 75.1\% at $N=158$ (\cref{tab:owm}).
\section{Analysis of Hindsight Supervision}\label{sec:when}

The reader need not see the entire execution, but it must observe a phase informative about the outcome. We first compare recording windows, then vary the amount and source of execution data. Across tasks, we compare the R2V gain with the gain from giving direct regression more labels.

\begin{figure*}[t]
\centering
\begin{tikzpicture}
\begin{groupplot}[group style={group size=3 by 1,horizontal sep=0.48cm},
  odplot,width=0.370\textwidth,height=4.6cm,
  title style={font=\small\bfseries,yshift=1pt},
  xlabel style={font=\scriptsize},ylabel style={font=\scriptsize},
  tick label style={font=\scriptsize},
]
\nextgroupplot[title={(a) Outcome visibility},
  xmin=0.57,xmax=1.0,xtick={0.6,0.7,0.8,0.9,1.0},
  ymin=-1.5,ymax=16,ytick={0,4,8,12,16},
  xlabel={Reader AUC},ylabel={R2V gain (points)},
  legend style={at={(0.98,0.04)},anchor=south east,font=\scriptsize,fill=white,draw=none,inner sep=1pt}]
\fill[cReadB!12] (axis cs:0.59,-1.5) rectangle (axis cs:0.73,16);
\addplot[only marks,cRead,mark=*,mark size=2pt] coordinates {(0.663,-0.2) (0.840,7.2) (0.848,9.0) (0.850,7.8) (0.945,9.3)};
\addlegendentry{$N=16$}
\addplot[only marks,cDirect,mark=square*,mark size=1.8pt] coordinates {(0.696,-0.4) (0.849,9.9) (0.851,11.5) (0.874,10.8) (0.971,13.6)};
\addlegendentry{$N=32$}
\node[font=\scriptsize,anchor=north,align=center,text=cGray] at (axis cs:0.66,15.4) {pre-action\\AUC range};
\node[font=\scriptsize,anchor=south] at (axis cs:0.68,0.6) {approach};
\node[font=\scriptsize,anchor=south] at (axis cs:0.84,11.8) {q25, close, q75};
\node[font=\scriptsize,anchor=east] at (axis cs:0.97,14.4) {late};
\nextgroupplot[title={(b) Recorded experience},
  xmode=log,log basis x=2,xmin=120,xmax=1280,
  xtick={142,270,540,1080},xticklabels={142,270,540,1080},
  ymin=0,ymax=18,ytick={0,4,8,12,16},xlabel={Unlabeled scenes},
  legend columns=2,legend style={at={(0.5,0.99)},anchor=north,font=\scriptsize,fill=white,draw=none,inner sep=1pt,column sep=1pt}]
\addplot[cRead,thick,mark=*,mark size=1.5pt] coordinates {(142,3.3) (270,4.2) (540,6.2) (1080,8.0)};
\addlegendentry{random, 16}
\addplot[cRead,thick,dashed,mark=*,mark size=1.5pt] coordinates {(142,5.6) (270,6.9) (540,11.0) (1080,10.9)};
\addlegendentry{random, 32}
\addplot[cDirect,thick,mark=square*,mark size=1.3pt] coordinates {(142,0.6) (270,1.0) (540,3.5) (1080,4.5)};
\addlegendentry{logged, 16}
\addplot[cDirect,thick,dashed,mark=square*,mark size=1.3pt] coordinates {(142,2.8) (270,5.8) (540,7.2) (1080,7.9)};
\addlegendentry{logged, 32}
\nextgroupplot[title={(c) Across tasks},
  xmin=-0.5,xmax=12,ymin=-0.8,ymax=12,
  xtick={0,4,8,12},ytick={0,4,8,12},
  xlabel={Label-limited headroom (points)}]
\addplot[cGray!60,dashed,domain=0:12] {x};
\addplot[only marks,cRead,mark=*,mark size=2.2pt] coordinates {(10.9,10.2) (4.7,4.1) (6.6,4.9) (1.3,1.3) (0.6,0.4)};
\node[font=\scriptsize,anchor=east] at (axis cs:10.6,10.3) {grasping};
\node[font=\scriptsize,anchor=south west] at (axis cs:6.6,4.9) {adjust bottle};
\node[font=\scriptsize,anchor=east] at (axis cs:4.5,4.0) {place can};
\node[font=\scriptsize,anchor=west] at (axis cs:1.5,1.6) {stamp};
\node[font=\scriptsize,anchor=north west] at (axis cs:0.8,0.4) {handover};
\end{groupplot}
\end{tikzpicture}
\caption{\textbf{Reader windows, recording budgets, and task-level gains.} (a) Reader AUC and verifier gain for different windows of the same grasp executions. (b) Verifier gain versus the number of new scenes, with one recorded action per scene. Panels (a,b) use neural screening results. (c) Confirmation gains across five tasks compare R2V at 16 labels with the improvement direct regression obtains from the full label set. The diagonal denotes equal gains.}
\label{fig:mechanism}
\end{figure*}
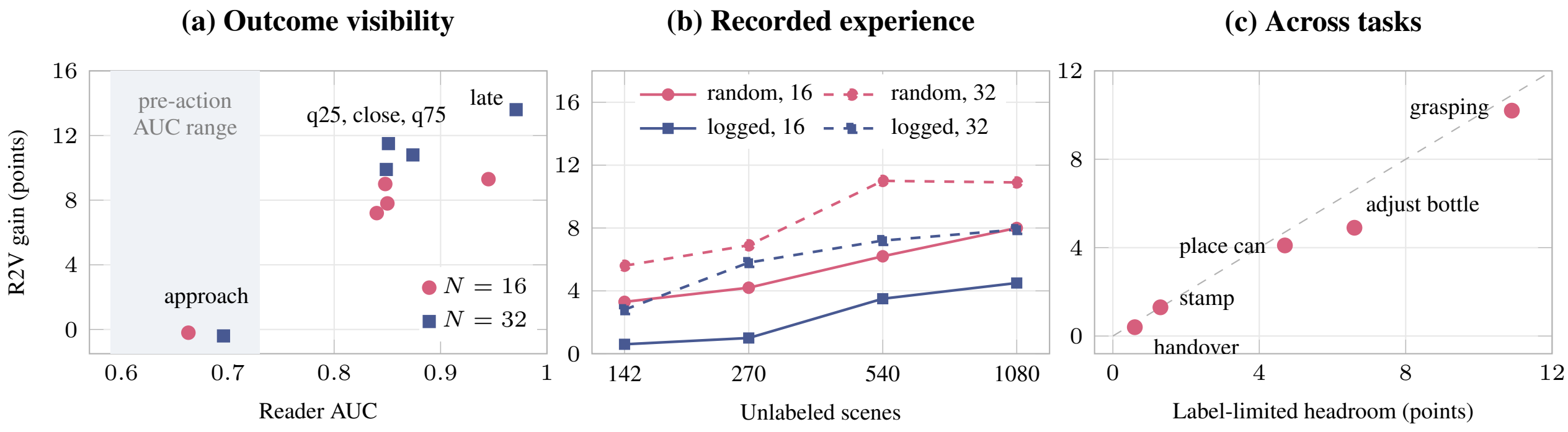

\subsection{Recording windows}
We extract several 1.2-second windows from each grasp execution. The approach window includes first contact but ends before gripper closure. Its reader reaches AUC 0.66--0.70 and produces little selection gain. Later windows, from q25 onward, give reader AUC of at least 0.84 and screening gains of 7.2--13.6 points over direct regression (\cref{fig:mechanism}a). On confirmation scenes, the close window improves success over approach by 8.4 and 9.5 points at $N=16,32$. These comparisons use the same executions, so the change is in which phase the reader observes.

\subsection{Single-execution recordings and deployment logs}
We collect one action per new scene, either at random or using a deployed selector. With 16 labeled episodes, increasing random recordings from 142 to 1,080 raises the screening gain from 3.3 to 8.0 points (\cref{fig:mechanism}b). We also train a verifier on recordings from the deployed selector. One logged execution from each of 540 new scenes improves confirmation success over direct regression by 5.5 points at $N=16$ and 7.1 at $N=32$ (\cref{tab:record}).

Random actions cover alternatives that the deployed selector may rarely choose. The logged data instead concentrate on that selector's preferred plans, and produce smaller gains per scene in the budget study. The full curves for both collection protocols appear in \cref{app:record}.

\begin{table*}[t]
\centering\small
\setlength{\tabcolsep}{5pt}
\caption{\textbf{Selection success on four additional manipulation tasks.} Confirmation success (\%) at $N=16$. Fixed order uses the same labeled scenes. Neural paired intervals resample scenes and label subsets. Gains are computed before rounding. The final column gives R2V gains under a matched tree learner. \Cref{app:tasks} gives all budgets and headrooms.}
\label{tab:taskgains}
\begin{tabular}{@{}lrrrrcr@{}}
\toprule
Task & Scenes & Fixed & Neural LR & Neural R2V & Neural gain [95\% CI] & Tree gain \\
\midrule
Place can basket & 400 & 74.1 & 69.8 & \cellcolor{ours}73.8 & $+4.1$ [1.9, 6.4] & $+3.1$ \\
Handover block & 400 & 95.3 & 96.4 & \cellcolor{ours}96.8 & $+0.4$ [0.0, 0.9] & $-0.3$ \\
Stamp seal & 400 & 66.7 & 64.3 & \cellcolor{ours}65.6 & $+1.3$ [0.5, 2.3] & $+1.6$ \\
Adjust bottle & 996 & 61.8 & 58.1 & \cellcolor{ours}63.0 & $+4.9$ [2.8, 7.3] & $+10.8$ \\
\bottomrule
\end{tabular}
\end{table*}

\subsection{Label and recording budgets}
\Cref{tab:grid} crosses the number of labeled episodes with the number of new scenes contributing one random execution each. Every entry is evaluated on the same 512 confirmation scenes. At $N=16$, neural success rises from 63.4\% with 142 recordings to 68.2\% with 1,080. The latter is close to the 68.4\% achieved by direct neural regression on 128 labeled episodes.

For trees, 16 labeled episodes and 1,080 recordings give 74.8\%, compared with 69.9\% for direct regression. The tree recording-budget curve is flatter than the neural curve, and increasing the label budget improves both.

\paragraph{Execution counts.}
A source episode can contain several candidate executions, so \cref{app:executions} also counts individual attempts. R2V uses 122 labeled executions and 1,071 unlabeled recordings to reach 70.6\% on the screening set. Direct regression on 2,797 labeled executions reaches 70.3\%. In this comparison, R2V uses fewer than half as many training executions as well as fewer outcome labels. Development data are identified separately.

\begin{figure}[t]
\centering
\begin{tikzpicture}
\begin{axis}[
 width=0.98\columnwidth,height=1.78in,
 tick label style={font=\scriptsize},label style={font=\small},
 legend cell align=left,grid=major,grid style={gray!18},
 axis line style={gray!60},tick style={gray!60},
 xlabel={Candidates $K$},ylabel={Success (\%)},
 ymin=35,ymax=76,ytick={40,50,60,70},
 xmin=0.8,xmax=6.2,xtick={1,2,3,4,5,6},xticklabels={1,2,4,8,16,all},
 legend style={at={(0.5,1.04)},anchor=south,legend columns=2,font=\scriptsize,draw=none,fill=none,column sep=4pt},
]
\addplot+[cDirect,mark=*,solid,line width=0.9pt,mark size=1.5pt] coordinates {(1,47.470085470) (2,54.415726496) (3,58.204444444) (4,59.552820513) (5,59.530940171) (6,59.623931624)};
\addlegendentry{Direct labels}
\addplot+[cReadB,mark=square*,densely dashed,line width=0.9pt,mark size=1.5pt] coordinates {(1,47.470085470) (2,56.644102564) (3,62.162735043) (4,65.011282051) (5,66.402051282) (6,66.666666667)};
\addlegendentry{141 recordings}
\addplot+[cGray,mark=triangle*,dotted,line width=0.9pt,mark size=1.5pt] coordinates {(1,47.470085470) (2,56.683076923) (3,61.830427350) (4,65.215726496) (5,66.962051282) (6,67.316239316)};
\addlegendentry{537 recordings}
\addplot+[cRead,mark=*,solid,line width=0.9pt,mark size=1.5pt] coordinates {(1,47.470085470) (2,57.712820513) (3,63.719658120) (4,67.327179487) (5,68.909401709) (6,69.230769231)};
\addlegendentry{1,076 recordings}
\end{axis}
\end{tikzpicture}
\caption{\textbf{Recording budget and candidate count.} Tree verifiers trained with 16 labeled episodes and one random execution per additional scene, evaluated on the same 195-scene grasping set. Legend counts are the available recordings: 141, 537, and 1,076. Direct labels uses no additional recordings. These training sets differ from the all-feasible pool in \cref{fig:tts}. Full tables and paired intervals are in \cref{app:recordingk}.}
\label{fig:recordingk}
\end{figure}
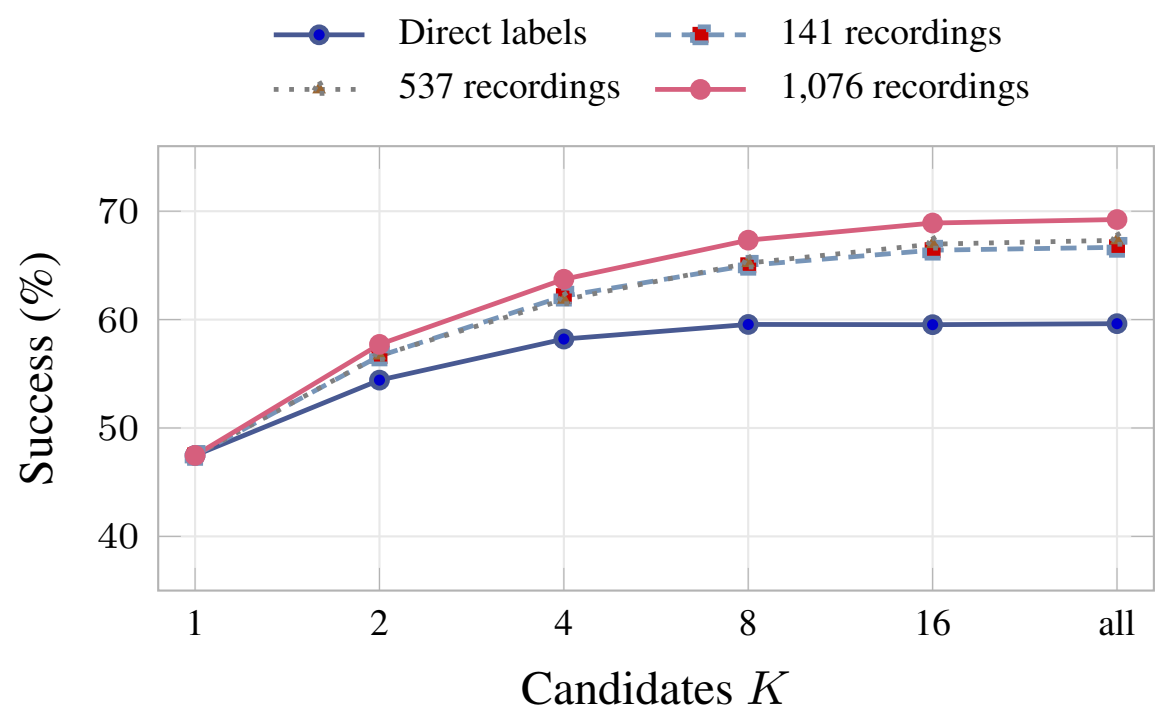

\paragraph{Recording budget and candidate count.}
We evaluate tree verifiers trained on nested single-execution recording sets, with 16 labeled episodes (\cref{fig:recordingk}). At $K=16$, the largest set improves success from 66.40\% to 68.91\% relative to the smallest, a paired gain of 2.51 points [0.44, 4.64]. Increasing $K$ from 4 to 16 gives 5.19 points for the largest set and 1.33 for direct regression. The complete grid, actual recording counts, and recording-by-$K$ interaction are reported in \cref{app:recordingk}.

\subsection{Additional manipulation tasks}
\Cref{tab:taskgains} reports contact and timing choices in RoboTwin and DOMINO. We compare the R2V gain at 16 labeled scenes with \emph{label-limited headroom}, the increase in direct-regression success from 16 to all training labels. This reference separates tasks where the direct learner is still label-limited from those where more labels change little.

At 16 labeled scenes, neural R2V gains 4.1 points on place\_can\_basket and 4.9 on adjust\_bottle, against headrooms of 4.7 and 6.6. Handover\_block and stamp\_seal have headrooms of 0.6 and 1.3 and gains of 0.4 and 1.3. For trees, place\_can\_basket and adjust\_bottle gain 3.1 and 10.8 points, compared with headrooms of 3.5 and 11.4 (\cref{fig:mechanism}c).

Before evaluation, we predicted task-level gains from the reader's advantage over pre-action prediction and the label-limited headroom on training scenes. Three of four predictions held. Place\_can\_basket was the exception. A single execution per training candidate suggested little room for improvement, while repeated executions gave larger headroom estimates. Outcomes changed across repeats for 34\% of its candidates (\cref{app:tasks}).

A fixed order learned from the labeled scenes remains competitive in these scripted pools, where a candidate index represents the same contact or timing choice across scenes. In dynamic grasping, the trajectories are replanned for each scene and do not have that reusable index meaning.

\subsection{Generated-future diagnostics}\label{sec:generated}
Instead of predicting a reader score, a selector can generate future observations for each action and score those observations. We compare the resulting scores with readouts of actual execution recordings.

\paragraph{PokeWorld pushing.}
Each of 48 scenes has four candidate pushes. A distance-based readout selects the best push from simulator-rendered futures in every scene. On eight DreamDojo samples per push, its top-1 accuracy is 27.1\% [14.6, 39.6], compared with 25\% chance, and its Spearman correlation is 0.21. The position extractor is accurate on the observed initial frame, with a maximum error of 0.8 pixels. Candidate rankings from the generated outcomes are much less accurate (\cref{app:push}).

\begin{table}[t]
\centering\footnotesize
\setlength{\tabcolsep}{2.6pt}
\caption{\textbf{Success discrimination in world-action-model rollouts.} Within-scene AUC for rollout success. The value and imagined-future scores are available at the selected decision. Recorded scores use terminal observations. Linear probes are fitted separately to imagined and recorded features using the same encoder and scene splits.}
\label{tab:wam}
\begin{tabular}{@{}lccc@{}}
\toprule
Model / task & Value & Imagined & Recorded \\
\midrule
Cosmos / PnPCounterToCab & 0.63 & 0.73 & 1.00 \\
Cosmos / TurnOffStove & 0.50 & 0.60 & 1.00 \\
Cosmos / OpenSingleDoor & 0.45 & 0.62 & 1.00 \\
Cosmos / PnPCounterToSink & 0.73 & 0.40 & 0.92 \\
Fast-WAM (joint) / open\_microwave & -- & 0.35 & 1.00 \\
\bottomrule
\end{tabular}
\end{table}

\paragraph{World action models.}
For Cosmos Policy, linear probes evaluated by scene-level leave-one-out obtain within-scene AUC 0.92--1.00 on terminal recordings across four RoboCasa~\cite{nasiriany2024robocasa} tasks. Probes fitted to imagined-future features give AUC 0.40--0.73 at the first gripper-closing decision, while the policy's value head gives 0.45--0.73. Fast-WAM on open\_microwave, with futures imagined through its joint video-action inference, gives AUC 1.00 from recorded outcomes and 0.35 from imagined observations at first contact (\cref{tab:wam}).

The Cosmos predictions target the end of the action chunk, whereas the recorded clips show the episode's terminal outcome. These are different prediction horizons. The comparison measures how well the available scores distinguish rollout outcomes at those points, rather than isolating video fidelity at a common horizon. \Cref{app:cosmos} gives the timing and recording windows.

\section{Discussion}\label{sec:discussion}
Random-action recordings yield larger improvements per scene than logs from the deployed selector, which mostly executes candidates it already prefers. This suggests collecting a broader range of actions for the next verifier. Applying reader scores only to the small labeled set gives no improvement, so the gain comes from the additional executions.

\paragraph{Closed-loop extension.}
A preliminary extension to repeated action-chunk selection with a frozen Diffusion Policy~\cite{chi2023diffusion} on PushT raises success from 42\% to 51\% at $K=16$, with no gain at smaller budgets (\cref{app:closedloop}).

\paragraph{Limitations.}
All experiments are in simulation, and physical-robot performance remains untested. The main experiments make one selection per scene before execution, and the closed-loop extension uses a single frozen policy. Reader observations must reveal the outcome.

\section{Conclusion}\label{sec:conclusion}
R2V trains a pre-action verifier on scores inferred from completed executions. It improves both tree and neural action selection over direct regression, with larger candidate pools yielding further gains in grasping. The benefit depends on observing the outcome and scoring executions beyond the labeled set, including one-action-per-scene logs.

\clearpage
{
    \small
    \bibliographystyle{ieeenat_fullname}
    \bibliography{main}

\begin{thebibliography}{52}
\providecommand{\natexlab}[1]{#1}
\providecommand{\url}[1]{\texttt{#1}}
\expandafter\ifx\csname urlstyle\endcsname\relax
  \providecommand{\doi}[1]{doi: #1}\else
  \providecommand{\doi}{doi: \begingroup \urlstyle{rm}\Url}\fi

\bibitem[Agarwal et~al.(2021)Agarwal, Schwarzer, Castro, Courville, and
  Bellemare]{NEURIPS2021_f514cec8}
Rishabh Agarwal, Max Schwarzer, Pablo~Samuel Castro, Aaron Courville, and
  Marc~G. Bellemare.
\newblock Deep reinforcement learning at the edge of the statistical precipice.
\newblock In \emph{Advances in Neural Information Processing Systems
  (NeurIPS)}, pages 29304--29320, 2021.

\bibitem[Andrychowicz et~al.(2017)Andrychowicz, Wolski, Ray, Schneider, Fong,
  Welinder, McGrew, Tobin, Abbeel, and Zaremba]{andrychowicz2017her}
Marcin Andrychowicz, Filip Wolski, Alex Ray, Jonas Schneider, Rachel Fong,
  Peter Welinder, Bob McGrew, Josh Tobin, Pieter Abbeel, and Wojciech Zaremba.
\newblock {Hindsight Experience Replay}.
\newblock In \emph{Advances in Neural Information Processing Systems
  (NeurIPS)}, 2017.

\bibitem[Assran et~al.(2025)Assran, Bardes, Fan, Garrido, Howes, Komeili,
  Muckley, Rizvi, Roberts, Sinha, Zholus, Arnaud, Gejji, Martin, Hogan, Dugas,
  Bojanowski, Khalidov, Labatut, Massa, Szafraniec, Krishnakumar, Li, Ma,
  Chandar, Meier, LeCun, Rabbat, and Ballas]{assran2025vjepa2}
Mido Assran, Adrien Bardes, David Fan, Quentin Garrido, Russell Howes, Mojtaba
  Komeili, Matthew Muckley, Ammar Rizvi, Claire Roberts, Koustuv Sinha, Artem
  Zholus, Sergio Arnaud, Abha Gejji, Ada Martin, Francois~Robert Hogan, Daniel
  Dugas, Piotr Bojanowski, Vasil Khalidov, Patrick Labatut, Francisco Massa,
  Marc Szafraniec, Kapil Krishnakumar, Yong Li, Xiaodong Ma, Sarath Chandar,
  Franziska Meier, Yann LeCun, Michael Rabbat, and Nicolas Ballas.
\newblock {V-JEPA} 2: Self-supervised video models enable understanding,
  prediction and planning.
\newblock \emph{arXiv preprint arXiv:2506.09985}, 2025.

\bibitem[Chen et~al.(2020)Chen, Zhou, Koltun, and
  Kr{\"a}henb{\"u}hl]{chen2019cheating}
Dian Chen, Brady Zhou, Vladlen Koltun, and Philipp Kr{\"a}henb{\"u}hl.
\newblock Learning by cheating.
\newblock In \emph{Proceedings of the Conference on Robot Learning (CoRL)},
  pages 66--75, 2020.

\bibitem[Chen et~al.(2026)Chen, Chen, Chen, Cai, Liu, Li, Liang, Lin, Ge, Gu,
  Deng, Guo, Nian, Xie, Chen, Su, Xu, Liu, Hu, Gao, Wang, Liang, Qin, Yang,
  Luo, and Mu]{chen2025robotwin2}
Tianxing Chen, Zanxin Chen, Baijun Chen, Zijian Cai, Yibin Liu, Zixuan Li,
  Qiwei Liang, Xianliang Lin, Yiheng Ge, Zhenyu Gu, Weiliang Deng, Yubin Guo,
  Tian Nian, Xuanbing Xie, Qiangyu Chen, Kailun Su, Tianling Xu, Guodong Liu,
  Mengkang Hu, {Huan-Ang} Gao, Kaixuan Wang, Zhixuan Liang, Yusen Qin, Xiaokang
  Yang, Ping Luo, and Yao Mu.
\newblock {RoboTwin} 2.0: A scalable data generator and benchmark with strong
  domain randomization for robust bimanual robotic manipulation.
\newblock In \emph{Proceedings of the 43rd International Conference on Machine
  Learning (ICML)}, pages 13673--13699, 2026.

\bibitem[Chi et~al.(2025)Chi, Xu, Feng, Cousineau, Du, Burchfiel, Tedrake, and
  Song]{chi2023diffusion}
Cheng Chi, Zhenjia Xu, Siyuan Feng, Eric Cousineau, Yilun Du, Benjamin
  Burchfiel, Russ Tedrake, and Shuran Song.
\newblock {Diffusion Policy}: Visuomotor policy learning via action diffusion.
\newblock \emph{The International Journal of Robotics Research}, 44\penalty0
  (10--11):\penalty0 1684--1704, 2025.

\bibitem[de~Boer et~al.(2005)de~Boer, Kroese, Mannor, and
  Rubinstein]{deboer2005cem}
Pieter-Tjerk de Boer, Dirk~P. Kroese, Shie Mannor, and Reuven~Y. Rubinstein.
\newblock A tutorial on the cross-entropy method.
\newblock \emph{Annals of Operations Research}, 134\penalty0 (1):\penalty0
  19--67, 2005.

\bibitem[Du et~al.(2023{\natexlab{a}})Du, Konyushkova, Denil, Raju, Landon,
  Hill, de~Freitas, and Cabi]{du2023success}
Yuqing Du, Ksenia Konyushkova, Misha Denil, Akhil Raju, Jessica Landon, Felix
  Hill, Nando de Freitas, and Serkan Cabi.
\newblock Vision-language models as success detectors.
\newblock In \emph{Proceedings of the 2nd Conference on Lifelong Learning
  Agents (CoLLAs)}, pages 120--136, 2023{\natexlab{a}}.

\bibitem[Du et~al.(2023{\natexlab{b}})Du, Yang, Dai, Dai, Nachum, Tenenbaum,
  Schuurmans, and Abbeel]{du2023unipi}
Yilun Du, Mengjiao Yang, Bo Dai, Hanjun Dai, Ofir Nachum, Joshua~B. Tenenbaum,
  Dale Schuurmans, and Pieter Abbeel.
\newblock Learning universal policies via text-guided video generation.
\newblock In \emph{Advances in Neural Information Processing Systems
  (NeurIPS)}, 2023{\natexlab{b}}.

\bibitem[Ebert et~al.(2018)Ebert, Finn, Dasari, Xie, Lee, and
  Levine]{ebert2018visual}
Frederik Ebert, Chelsea Finn, Sudeep Dasari, Annie Xie, Alex Lee, and Sergey
  Levine.
\newblock Visual foresight: Model-based deep reinforcement learning for
  vision-based robotic control.
\newblock \emph{arXiv preprint arXiv:1812.00568}, 2018.

\bibitem[Escontrela et~al.(2023)Escontrela, Adeniji, Yan, Jain, Peng, Goldberg,
  Lee, Hafner, and Abbeel]{escontrela2023viper}
Alejandro Escontrela, Ademi Adeniji, Wilson Yan, Ajay Jain, Xue~Bin Peng, Ken
  Goldberg, Youngwoon Lee, Danijar Hafner, and Pieter Abbeel.
\newblock Video prediction models as rewards for reinforcement learning.
\newblock In \emph{Advances in Neural Information Processing Systems
  (NeurIPS)}, pages 68760--68783, 2023.

\bibitem[Fang et~al.(2026{\natexlab{a}})Fang, Li, Wang, Xi, Liang, and
  Bai]{fang2026domino}
Heng Fang, Shangru Li, Shuhan Wang, Xuanyang Xi, Dingkang Liang, and Xiang Bai.
\newblock Towards generalizable robotic manipulation in dynamic environments.
\newblock In \emph{European Conference on Computer Vision (ECCV)}, pages
  314--333, 2026{\natexlab{a}}.

\bibitem[Fang et~al.(2026{\natexlab{b}})Fang, Chen, and Cai]{fang2026pfd}
Pengcheng Fang, Hongli Chen, and Xiaohao Cai.
\newblock Privileged foresight distillation: Zero-cost future correction for
  world action models.
\newblock \emph{arXiv preprint arXiv:2604.25859}, 2026{\natexlab{b}}.

\bibitem[Farahmand et~al.(2017)Farahmand, Barreto, and
  Nikovski]{pmlr-v54-farahmand17a}
Amir-Massoud Farahmand, Andr{\'e} Barreto, and Daniel Nikovski.
\newblock Value-aware loss function for model-based reinforcement learning.
\newblock In \emph{Proceedings of the 20th International Conference on
  Artificial Intelligence and Statistics (AISTATS)}, pages 1486--1494, 2017.

\bibitem[Finn and Levine(2017)]{finn2017deep}
Chelsea Finn and Sergey Levine.
\newblock Deep visual foresight for planning robot motion.
\newblock In \emph{IEEE International Conference on Robotics and Automation
  (ICRA)}, pages 2786--2793, 2017.

\bibitem[Gao et~al.(2026)Gao, Liang, Zheng, Malik, Ye, Yu, Tseng, Dong, Mo,
  Lin, Xiang, Xie, Zheng, Niu, Jannaty, Gu, Zhang, Malik, Abbeel, Liu, Zhu,
  Jang, and Fan]{gao2026dreamdojo}
Shenyuan Gao, William Liang, Kaiyuan Zheng, Ayaan~Naveed Malik, Seonghyeon Ye,
  Sihyun Yu, Wei-Cheng Tseng, Yuzhu Dong, Kaichun Mo, Chen-Hsuan Lin, Jiannan
  Xiang, Yuqi Xie, Ruijie Zheng, Dantong Niu, Pooya Jannaty, Jinwei Gu, Jun
  Zhang, Jitendra Malik, Pieter Abbeel, Ming-Yu Liu, Yuke Zhu, Joel Jang, and
  Linxi Fan.
\newblock {DreamDojo}: A generalist robot world model from large-scale human
  videos.
\newblock In \emph{Proceedings of the 43rd International Conference on Machine
  Learning (ICML)}, pages 33340--33366, 2026.

\bibitem[Grimm et~al.(2020)Grimm, Barreto, Singh, and Silver]{grimm2020value}
Christopher Grimm, Andr{\'e} Barreto, Satinder Singh, and David Silver.
\newblock The value equivalence principle for model-based reinforcement
  learning.
\newblock In \emph{Advances in Neural Information Processing Systems
  (NeurIPS)}, pages 5541--5552, 2020.

\bibitem[Hinton et~al.(2015)Hinton, Vinyals, and Dean]{hinton2015distilling}
Geoffrey Hinton, Oriol Vinyals, and Jeff Dean.
\newblock Distilling the knowledge in a neural network.
\newblock \emph{arXiv preprint arXiv:1503.02531}, 2015.

\bibitem[Holm(1979)]{holm1979}
Sture Holm.
\newblock A simple sequentially rejective multiple test procedure.
\newblock \emph{Scandinavian Journal of Statistics}, 6\penalty0 (2):\penalty0
  65--70, 1979.

\bibitem[Hsieh et~al.(2026)Hsieh, Zheng, Chen, Chan, and
  Darrell]{hsieh2026verification}
Wen-Han Hsieh, Ken~Jiankun Zheng, Chia-Chien Chen, David~M. Chan, and Trevor
  Darrell.
\newblock Learning low-latency action verification for frozen
  vision-language-action models.
\newblock \emph{arXiv preprint arXiv:2611.14834}, 2026.

\bibitem[Kalashnikov et~al.(2018)Kalashnikov, Irpan, Pastor, Ibarz, Herzog,
  Jang, Quillen, Holly, Kalakrishnan, Vanhoucke, and
  Levine]{kalashnikov2018qtopt}
Dmitry Kalashnikov, Alex Irpan, Peter Pastor, Julian Ibarz, Alexander Herzog,
  Eric Jang, Deirdre Quillen, Ethan Holly, Mrinal Kalakrishnan, Vincent
  Vanhoucke, and Sergey Levine.
\newblock Scalable deep reinforcement learning for vision-based robotic
  manipulation.
\newblock In \emph{Proceedings of the 2nd Conference on Robot Learning (CoRL)},
  pages 651--673, 2018.

\bibitem[Kim et~al.(2026)Kim, Gao, Lin, et~al.]{kim2026cosmospolicy}
Moo~Jin Kim, Yihuai Gao, Tsung-Yi Lin, et~al.
\newblock {Cosmos Policy}: Fine-tuning video models for visuomotor control and
  planning.
\newblock \emph{arXiv preprint arXiv:2601.16163}, 2026.

\bibitem[Krzywicki et~al.(2023)Krzywicki, Ciebiera, Michaluk, Maziarz, and
  Cygan]{krzywicki2023grasping}
Piotr Krzywicki, Krzysztof Ciebiera, Rafa{\l} Michaluk, Inga Maziarz, and Marek
  Cygan.
\newblock Grasping student: Semi-supervised learning for robotic manipulation.
\newblock \emph{arXiv preprint arXiv:2303.04452}, 2023.

\bibitem[Kwok et~al.(2025)Kwok, Agia, Sinha, Foutter, Li, Stoica, Mirhoseini,
  and Pavone]{kwok2025robomonkey}
Jacky Kwok, Christopher Agia, Rohan Sinha, Matt Foutter, Shulu Li, Ion Stoica,
  Azalia Mirhoseini, and Marco Pavone.
\newblock {RoboMonkey}: Scaling test-time sampling and verification for
  vision-language-action models.
\newblock In \emph{Proceedings of the 9th Conference on Robot Learning (CoRL)},
  pages 3200--3217, 2025.

\bibitem[Lambert et~al.(2020)Lambert, Amos, Yadan, and
  Calandra]{lambert2020objective}
Nathan Lambert, Brandon Amos, Omry Yadan, and Roberto Calandra.
\newblock Objective mismatch in model-based reinforcement learning.
\newblock In \emph{Proceedings of the 2nd Conference on Learning for Dynamics
  and Control (L4DC)}, pages 761--770, 2020.

\bibitem[Lee et~al.(2026)Lee, Wagenmaker, Pertsch, Liang, Levine, and
  Finn]{lee2026roboreward}
Tony Lee, Andrew Wagenmaker, Karl Pertsch, Percy Liang, Sergey Levine, and
  Chelsea Finn.
\newblock {RoboReward}: General-purpose vision-language reward models for
  robotics.
\newblock \emph{arXiv preprint arXiv:2601.00675}, 2026.

\bibitem[Levine et~al.(2016)Levine, Pastor, Krizhevsky, and
  Quillen]{levine2016handeye}
Sergey Levine, Peter Pastor, Alex Krizhevsky, and Deirdre Quillen.
\newblock Learning hand-eye coordination for robotic grasping with deep
  learning and large-scale data collection.
\newblock \emph{arXiv preprint arXiv:1603.02199}, 2016.

\bibitem[Li et~al.(2025)Li, Gao, Sadigh, and Song]{li2025uva}
Shuang Li, Yihuai Gao, Dorsa Sadigh, and Shuran Song.
\newblock Unified video action model.
\newblock In \emph{Robotics: Science and Systems (RSS)}, 2025.

\bibitem[Lopez-Paz et~al.(2016)Lopez-Paz, Bottou, Sch{\"o}lkopf, and
  Vapnik]{lopezpaz2016unifying}
David Lopez-Paz, L{\'e}on Bottou, Bernhard Sch{\"o}lkopf, and Vladimir Vapnik.
\newblock Unifying distillation and privileged information.
\newblock In \emph{International Conference on Learning Representations
  (ICLR)}, 2016.

\bibitem[Luo et~al.(2025)Luo, Xu, Wu, and Levine]{luo2024hilserl}
Jianlan Luo, Charles Xu, Jeffrey Wu, and Sergey Levine.
\newblock Precise and dexterous robotic manipulation via human-in-the-loop
  reinforcement learning.
\newblock \emph{Science Robotics}, 10\penalty0 (105):\penalty0 eads5033, 2025.

\bibitem[Ma et~al.(2023)Ma, Sodhani, Jayaraman, Bastani, Kumar, and
  Zhang]{ma2023vip}
Yecheng~Jason Ma, Shagun Sodhani, Dinesh Jayaraman, Osbert Bastani, Vikash
  Kumar, and Amy Zhang.
\newblock {VIP}: Towards universal visual reward and representation via
  value-implicit pre-training.
\newblock In \emph{International Conference on Learning Representations
  (ICLR)}, 2023.

\bibitem[Mur-Labadia et~al.(2026)Mur-Labadia, Muckley, Bar, Assran, Sinha,
  Rabbat, LeCun, Ballas, and Bardes]{murlabadia2026vjepa21}
Lorenzo Mur-Labadia, Matthew Muckley, Amir Bar, Mido Assran, Koustuv Sinha,
  Mike Rabbat, Yann LeCun, Nicolas Ballas, and Adrien Bardes.
\newblock {V-JEPA} 2.1: Unlocking dense features in video self-supervised
  learning.
\newblock \emph{arXiv preprint arXiv:2603.14482}, 2026.

\bibitem[Nakamoto et~al.(2025)Nakamoto, Mees, Kumar, and
  Levine]{pmlr-v270-nakamoto25a}
Mitsuhiko Nakamoto, Oier Mees, Aviral Kumar, and Sergey Levine.
\newblock Steering your generalists: Improving robotic foundation models via
  value guidance.
\newblock In \emph{Proceedings of the 8th Conference on Robot Learning (CoRL)},
  pages 4996--5013, 2025.

\bibitem[Nasiriany et~al.(2024)Nasiriany, Maddukuri, Zhang, Parikh, Lo, Joshi,
  Mandlekar, and Zhu]{nasiriany2024robocasa}
Soroush Nasiriany, Abhiram Maddukuri, Lance Zhang, Adeet Parikh, Aaron Lo,
  Abhishek Joshi, Ajay Mandlekar, and Yuke Zhu.
\newblock {RoboCasa}: Large-scale simulation of everyday tasks for generalist
  robots.
\newblock \emph{arXiv preprint arXiv:2406.02523}, 2024.

\bibitem[{NVIDIA}(2025)]{nvidia2025cosmosreason2}
{NVIDIA}.
\newblock {Cosmos-Reason2}: Physical {AI} common sense and embodied reasoning
  models.
\newblock \url{https://huggingface.co/nvidia/Cosmos-Reason2-8B}, 2025.
\newblock Model card for Cosmos-Reason2-8B, released December 19, 2025.

\bibitem[Oquab et~al.(2024)Oquab, Darcet, Moutakanni, Vo, Szafraniec, Khalidov,
  Fernandez, Haziza, Massa, El-Nouby, Assran, Ballas, Galuba, Howes, Huang, Li,
  Misra, Rabbat, Sharma, Synnaeve, Xu, Jegou, Mairal, Labatut, Joulin, and
  Bojanowski]{oquab2024dinov2}
Maxime Oquab, Timoth{\'e}e Darcet, Th{\'e}o Moutakanni, Huy Vo, Marc
  Szafraniec, Vasil Khalidov, Pierre Fernandez, Daniel Haziza, Francisco Massa,
  Alaaeldin El-Nouby, Mahmoud Assran, Nicolas Ballas, Wojciech Galuba, Russell
  Howes, Po-Yao Huang, Shang-Wen Li, Ishan Misra, Michael Rabbat, Vasu Sharma,
  Gabriel Synnaeve, Hu Xu, Herv{\'e} Jegou, Julien Mairal, Patrick Labatut,
  Armand Joulin, and Piotr Bojanowski.
\newblock {DINOv2}: Learning robust visual features without supervision.
\newblock \emph{Transactions on Machine Learning Research}, 2024.

\bibitem[{Physical Intelligence} et~al.(2025)]{pi2025pistar}
{Physical Intelligence} et~al.
\newblock $\pi^{*}_{0.6}$: a {VLA} that learns from experience.
\newblock \emph{arXiv preprint arXiv:2511.14759}, 2025.

\bibitem[Sch{\"o}lkopf et~al.(2012)Sch{\"o}lkopf, Janzing, Peters, Sgouritsa,
  Zhang, and Mooij]{scholkopf2012causal}
Bernhard Sch{\"o}lkopf, Dominik Janzing, Jonas Peters, Eleni Sgouritsa, Kun
  Zhang, and Joris Mooij.
\newblock On causal and anticausal learning.
\newblock In \emph{Proceedings of the 29th International Conference on Machine
  Learning (ICML)}, 2012.

\bibitem[Snell et~al.(2025)Snell, Lee, Xu, and Kumar]{snell2024scaling}
Charlie Snell, Jaehoon Lee, Kelvin Xu, and Aviral Kumar.
\newblock Scaling {LLM} test-time compute optimally can be more effective than
  scaling parameters for reasoning.
\newblock In \emph{International Conference on Learning Representations
  (ICLR)}, 2025.

\bibitem[Song et~al.(2025)Song, Qu, Yao, Chen, Lv, Tang, Shi, Ren, Yao, Zhao,
  Wang, and Li]{song2025hume}
Haoming Song, Delin Qu, Yuanqi Yao, Qizhi Chen, Qi Lv, Yiwen Tang, Modi Shi,
  Guanghui Ren, Maoqing Yao, Bin Zhao, Dong Wang, and Xuelong Li.
\newblock Hume: Introducing system-2 thinking in visual-language-action model.
\newblock \emph{arXiv preprint arXiv:2505.21432}, 2025.

\bibitem[Sundaralingam et~al.(2023)Sundaralingam, Hari, Fishman, Garrett, Wyk,
  Blukis, Millane, Oleynikova, Handa, Ramos, Ratliff, and
  Fox]{sundaralingam2023curobo}
Balakumar Sundaralingam, Siva Kumar~Sastry Hari, Adam Fishman, Caelan Garrett,
  Karl~Van Wyk, Valts Blukis, Alexander Millane, Helen Oleynikova, Ankur Handa,
  Fabio Ramos, Nathan Ratliff, and Dieter Fox.
\newblock {cuRobo}: Parallelized collision-free minimum-jerk robot motion
  generation.
\newblock \emph{arXiv preprint arXiv:2310.17274}, 2023.

\bibitem[Tan et~al.(2026)Tan, Xu, Xu, Tao, Li, Hong, Feng, Du, and
  Wang]{tan2026identifiability}
Kaizhen Tan, Sizhe Xu, Xin Xu, Siru Tao, Yixiao Li, Hanzhe Hong, Yang Feng,
  Heqing Du, and Zhaonan Wang.
\newblock What can latent world models know? {P}hysical information in
  multimodal predictive representations.
\newblock \emph{arXiv preprint arXiv:2607.27017}, 2026.

\bibitem[Vapnik and Vashist(2009)]{vapnik2009lupi}
Vladimir Vapnik and Akshay Vashist.
\newblock A new learning paradigm: Learning using privileged information.
\newblock \emph{Neural Networks}, 22:\penalty0 544--557, 2009.

\bibitem[Williams et~al.(2017)Williams, Wagener, Goldfain, Drews, Rehg, Boots,
  and Theodorou]{williams2017mppi}
Grady Williams, Nolan Wagener, Brian Goldfain, Paul Drews, James~M. Rehg, Byron
  Boots, and Evangelos~A. Theodorou.
\newblock Information theoretic {MPC} for model-based reinforcement learning.
\newblock In \emph{IEEE International Conference on Robotics and Automation
  (ICRA)}, pages 1714--1721, 2017.

\bibitem[Xie et~al.(2020)Xie, Luong, Hovy, and Le]{xie2020noisy}
Qizhe Xie, Minh-Thang Luong, Eduard Hovy, and Quoc~V. Le.
\newblock Self-training with {Noisy Student} improves {ImageNet}
  classification.
\newblock In \emph{Proceedings of the IEEE/CVF Conference on Computer Vision
  and Pattern Recognition (CVPR)}, pages 10687--10698, 2020.

\bibitem[Xu et~al.(2026)Xu, Jiang, Marangola, Agrawal, and Hong]{xu2026lfh}
Iris Xu, Sunshine Jiang, John Marangola, Pulkit Agrawal, and Zhang-Wei Hong.
\newblock Learning from hindsight for {VLA} reinforcement learning.
\newblock \emph{arXiv preprint arXiv:2607.09042}, 2026.

\bibitem[Ye et~al.(2026)Ye, Ge, Zheng, et~al.]{ye2026dreamzero}
Seonghyeon Ye, Yunhao Ge, Kaiyuan Zheng, et~al.
\newblock World action models are zero-shot policies.
\newblock \emph{arXiv preprint arXiv:2602.15922}, 2026.

\bibitem[Yuan et~al.(2026)Yuan, Dong, Liu, and Zhao]{yuan2026fastwam}
Tianyuan Yuan, Zibin Dong, Yicheng Liu, and Hang Zhao.
\newblock {Fast-WAM}: Do world action models need test-time future imagination?
\newblock \emph{arXiv preprint arXiv:2603.16666}, 2026.

\bibitem[Zhai et~al.(2023)Zhai, Mustafa, Kolesnikov, and Beyer]{zhai2023siglip}
Xiaohua Zhai, Basil Mustafa, Alexander Kolesnikov, and Lucas Beyer.
\newblock Sigmoid loss for language image pre-training.
\newblock In \emph{Proceedings of the IEEE/CVF International Conference on
  Computer Vision (ICCV)}, pages 11975--11986, 2023.

\bibitem[Zhang et~al.(2026)Zhang, Ito, Hoshino, Ikehata, and
  Sato]{zhang2026wcd}
Chuhan Zhang, Seiji Ito, Kenta Hoshino, Satoshi Ikehata, and Ikuro Sato.
\newblock World-coherent decoding: Self-verifying test-time planning for world
  action models.
\newblock \emph{arXiv preprint arXiv:2609.02159}, 2026.

\bibitem[Zhao et~al.(2026)Zhao, Cho, Shen, Zheng, Gao, Cao, and
  Mao]{zhao2026geobon}
Zesen Zhao, Minkyoung Cho, Hui Shen, Boyuan Zheng, Kunxiao Gao, Yulong Cao, and
  Z.~Morley Mao.
\newblock Test-time scaling for world action models via zero-shot geometric
  evaluation.
\newblock \emph{arXiv preprint arXiv:2607.17454}, 2026.

\bibitem[Zhou et~al.(2025)Zhou, Pan, LeCun, and Pinto]{zhou2025dinowm}
Gaoyue Zhou, Hengkai Pan, Yann LeCun, and Lerrel Pinto.
\newblock {DINO-WM}: World models on pre-trained visual features enable
  zero-shot planning.
\newblock In \emph{Proceedings of the 42nd International Conference on Machine
  Learning (ICML)}, pages 79115--79135, 2025.

\end{thebibliography}
}

\clearpage
\setcounter{page}{1}
\renewcommand{\thepage}{S\arabic{page}}
\appendix
\crefalias{section}{appendix}\crefalias{subsection}{subappendix}
\crefname{appendix}{Appendix}{Appendices}\Crefname{appendix}{Appendix}{Appendices}
\crefname{subappendix}{Appendix}{Appendices}\Crefname{subappendix}{Appendix}{Appendices}
\renewcommand{\floatpagefraction}{0.9}
\renewcommand{\dblfloatpagefraction}{0.9}
\renewcommand{\topfraction}{0.95}
\renewcommand{\textfraction}{0.05}
\setcounter{topnumber}{4}
\setcounter{totalnumber}{6}

\lefthyphenmin=3\relax
\righthyphenmin=3\relax
\hyphenation{McNemar}

\section{Test-Time Use}\label{app:testtime}

\begin{figure*}[t]
\centering
\includegraphics[width=\linewidth]{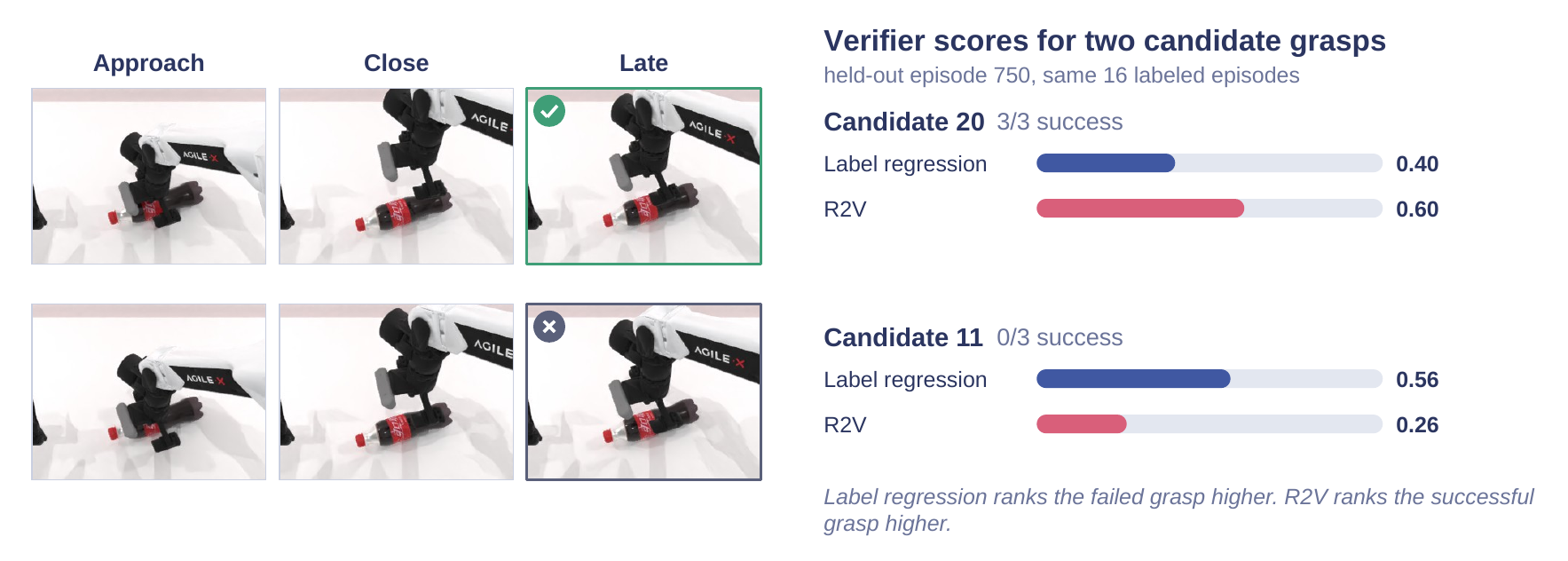}
\caption{\textbf{Candidate scores in held-out episode 750.} Frames show three windows from one execution of each candidate. Both tree verifiers use the same 16 labeled episodes. Candidate 20 succeeds in 3/3 executions and candidate 11 in 0/3. Display scores are monotonic transforms of each model's output, used to show its ordering.}
\label{fig:qual}
\end{figure*}

R2V scores candidates from the scene and action description and executes the highest-scoring feasible plan. The verifier is trained on reader scores. In the stored-candidate studies, the feasible-and-recorded mask includes a candidate only if it passes planning feasibility checks and has the stored outcomes required by that comparison. All selectors in a comparison use this same mask and the same executions.

\paragraph{Qualitative example.} The example in \cref{fig:qual} uses held-out seed 750 and label subset 0. Candidates 20 and 11 succeed in 3/3 and 0/3 executions. The pair was selected from a shortlist in which direct tree regression misordered the pair or separated its scores by less than 0.05, while tree R2V ordered it correctly with a gap of at least 0.2. These thresholds use the displayed scores $\sigma((s-\mu)/\tau)$, where $\mu$ and $\tau$ are each model's labeled-training score mean and standard deviation. This transform preserves ordering. Frames show the approach-window endpoint and close/late midpoints from the first of the three executions.

\section{Evaluation Protocol}\label{app:protocol}

The comparisons use stored scene motions and complete planned trajectories so that each selector chooses from the same candidate set. Executions are repeated because contact dynamics can vary even when a trajectory is held fixed. Across the 640 screening and confirmation episodes (seeds 512--1151), 6.9\% of 8,813 candidates changed outcome across their three executions.

The repeat-held-out best-of-$K$ reference in \cref*{fig:tts} selects with two executions and is evaluated on the third. The third execution independently evaluates this finite-sample selector. The population opportunity in \cref*{eq:selection} is defined using true success probabilities.

\paragraph{Labels, development episodes and epochs.}
The grasping fit pool contains 158 scenes from the training seed range and 2,797 executed candidates. Budgets $N\in\{16,32,64,128,158\}$ reveal outcomes for the selected scenes. Sixteen scenes contain about 280 labeled executions. Five labeled subsets (IDs 0--4) are shared across methods.

The main paper reports an average of 18.6 feasible planner proposals per scene. The fit pool contains 17.7 executed candidates per scene, while the repeated-execution set averages 13.8 recorded candidate trajectories per scene over its 640 screening and confirmation scenes, with three executions per trajectory. These are protocol-specific executed/recorded populations rather than alternative estimates of planner feasibility.

The 32 development scenes (seeds 160--191) measure reader AUC and are excluded from fitting the primary label-budget verifiers. Their outcomes were used for early stopping of the close-window feature predictor, fixed before the label-budget study, and for the separate external-validation comparisons in \cref{app:robust}. The close-window start time is computed from training plans (\cref{app:details}). Primary neural verifiers select epochs by four-fold cross-validation within the labeled training scenes.

\begin{table}[!tb]
\centering\footnotesize
\setlength{\tabcolsep}{3pt}
\caption{\textbf{Grasping and additional-task evaluation populations.} Each table and figure compares methods within its stated protocol.}
\label{tab:protocols}
\begin{tabular}{@{}p{0.34\linewidth}p{0.18\linewidth}p{0.40\linewidth}@{}}
\toprule
Experiment & Evaluation & Main distinction \\
\midrule
Reader AUC & 32 dev scenes & Recorded windows and pre-action prediction \\
Grasp K-curve & 195 confirm scenes & At least 16 feasible candidates per scene \\
Main selection & 512 confirm scenes & Complete confirmation population \\
Label-budget curve & 128 screen scenes & Neural curve in \cref{fig:budget} \\
Recording budgets & 128 / 512 scenes & Screening curves and confirmation in \cref{tab:grid,tab:record} \\
Recording by $K$ & 195 confirm scenes & Stored candidate draws in \cref{tab:recording_k_full} \\
Encoder comparison & 128 screen scenes & Re-encoded clips with matched learners \\
Target comparison & 512 confirm scenes & Neural epochs use 32 external dev scenes \\
Additional tasks & 400 / 996 confirm & Eight scripted candidates per scene \\
\bottomrule
\end{tabular}
\end{table}

\paragraph{Paired comparisons.} Shared candidate outcomes make agreement contribute zero to a paired difference. Screening and confirmation use separate scenes. Registered contrasts were fixed before confirmation evaluation. For the close-versus-approach controls, we average the selected candidate's three outcomes, then average over five label subsets within each episode. The test unit is one of the 512 evaluation episodes, not a repeat or label subset. A two-sided Monte Carlo sign-flip test uses 10,000 random sign vectors and the absolute mean paired difference as its statistic. The $p$ value is $(b+1)/(10{,}000+1)$, where $b$ counts resampled statistics at least as large as the observed statistic. The 95\% percentile intervals use 10,000 episode-bootstrap resamples. Both calculations use NumPy random seed 0.

\paragraph{Registered contrasts.} The daggered comparisons in main-paper \cref*{tab:main} are the registered paired comparisons reported there and use the sign-flip/episode-bootstrap procedure above with Holm correction over their prespecified family. Close minus approach at $N=16,32$ forms a two-contrast Holm family. The four outcome-feature comparisons in \cref{tab:owm} form a separate family: outcome world model minus label regression or R2V, each at $N=64,158$. Random and logged recordings in \cref{tab:record} each use a four-contrast family: R2V minus label regression or self-training, at $N=16,32$. The additional-task and synchronization families are specified in \cref{app:tasks,app:baselines}. PushT is reported separately in \cref{app:closedloop}.

\section{Windows, Reader and Training Details}\label{app:details}

\paragraph{Recorded windows.} Each window is a 1.2-second clip encoded by the frozen encoder into a 768-dimensional feature. Window start times are fixed for all episodes and computed from planned arrival times in the training scenes (seeds below 160): close starts at their median, q25 and q75 at their quartiles, and late 0.6~s after close. We use close as the default window and evaluate late as a separate timing ablation.

\begin{table}[!tb]
\centering\footnotesize
\setlength{\tabcolsep}{4pt}
\caption*{\textbf{Window values underlying main-paper \cref*{fig:mechanism}a.} AUC uses 32 development scenes. Verifier gains over direct regression use 128 screening scenes and five label subsets. Gains are in percentage points.}
\begin{tabular}{@{}lrrrr@{}}
\toprule
 & \multicolumn{2}{c}{$N=16$} & \multicolumn{2}{c}{$N=32$} \\
\cmidrule(lr){2-3}\cmidrule(l){4-5}
Window & AUC & Gain & AUC & Gain \\
\midrule
Approach & 0.663 & $-0.2$ & 0.696 & $-0.4$ \\
q25 & 0.840 & $+7.2$ & 0.849 & $+9.9$ \\
\rowcolor{ours} Close & 0.848 & $+9.0$ & 0.851 & $+11.5$ \\
q75 & 0.850 & $+7.8$ & 0.874 & $+10.8$ \\
Late & 0.945 & $+9.3$ & 0.971 & $+13.6$ \\
\bottomrule
\end{tabular}
\end{table}

\paragraph{Reader.} The reader is a logistic regression with a linear logit on close-window features, standardized with the mean and standard deviation of the labeled executions it is fit on. It is fit for 400 steps with an L2 coefficient of 0.001 on the weights and retrained for every label subset. The verifier regresses the reader's logit on every recorded execution, the labeled ones included, after standardizing it with the mean and standard deviation of the rows it is trained on.

\paragraph{Verifiers and controls.} R2V, label regression and self-training share the same two-hidden-layer MLP: Linear$(d,372)$--GELU--Dropout$(0.1)$--Linear$(372,372)$--GELU--Dropout$(0.1)$--Linear$(372,1)$. Training uses squared loss on standardized targets, AdamW (learning rate $10^{-3}$, weight decay $10^{-4}$, batch 64) and an ensemble of three seeds averaged at test time. Epochs are selected from up to 80 epochs by four-fold cross-validation within the available labeled episodes. The ensemble seeds are 0, 1, and 2. Self-training uses continuous predictions for unlabeled inputs and performs one refit. The outcome world model regresses the 768-dimensional window feature with the same MLP body and scores a candidate by applying the reader to the prediction.

\paragraph{Label subsets.} For each budget, episodes are sampled uniformly without replacement with subset seeds 0--4. Each method uses the same episode subset. At $N=158$, the five runs use every episode and different training randomness.

\paragraph{Encoders.} The primary video encoder is V-JEPA~2.1 ViT-B, using the publicly released ViT-B checkpoint distilled from ViT-G at $384$ resolution. For the comparison in \cref{tab:general}, we re-encode every clip with each encoder using the same pipeline. The pre-action scene feature and the reader's window feature come from the same encoder, and the plan summary, timing and commanded trajectory are unchanged. Readers, verifiers, label subsets and epoch selection follow the main runs.

\paragraph{Brier scores.} At $N=16$, the late, close and approach readers have Brier scores of 0.108, 0.194 and 0.313, against 0.247 for label regression. Brier score measures squared probability error and depends on both discrimination and calibration. The approach reader scores worse than the constant prediction 0.5 (Brier score 0.25) because its early clip carries little outcome evidence.

\section{Label Budgets and Training Variants}\label{app:budget}

\begin{figure}[t]
\centering
\begin{tikzpicture}
\begin{axis}[
  odplot,width=\linewidth,height=4.5cm,
  xmin=-0.1,xmax=4.1,ymin=55,ymax=81,
  xtick={0,1,2,3,4},xticklabels={16,32,64,128,158},
  ytick={55,60,65,70,75,80},
  xlabel={Labeled episodes $N$},ylabel={Success (\%)},
  legend columns=2,legend style={at={(0.5,1.05)},anchor=south,
  font=\scriptsize,draw=none,fill=none,column sep=3pt},
  tick label style={font=\scriptsize},label style={font=\footnotesize},
]
\addplot[cReadB,mark=triangle*,thick,mark size=1.6pt] coordinates {(0,68.1) (1,70.5) (2,74.3) (3,76) (4,77.6)};
\addlegendentry{Outcome world model}
\addplot[cRead,mark=*,thick,mark size=1.6pt] coordinates {(0,70.6) (1,73.6) (2,74) (3,75.1) (4,74.7)};
\addlegendentry{R2V}
\addplot[cDirect,mark=diamond*,thick,mark size=1.6pt] coordinates {(0,61.6) (1,62.1) (2,67.4) (3,69.6) (4,70.3)};
\addlegendentry{Label regression}
\addplot[cSelf,dashed,mark=square*,thick,mark size=1.6pt] coordinates {(0,58.9) (1,61) (2,65.8) (3,67.9) (4,70.3)};
\addlegendentry{Self-training}
\end{axis}
\end{tikzpicture}
\caption{\textbf{Label-budget scaling.} Neural success on 128 screening scenes, using the same labeled training episodes. Outcome-feature epochs use 32 external development scenes. Other models use within-training cross-validation. At $N=158$, all training executions are labeled.}
\label{fig:budget}
\end{figure}
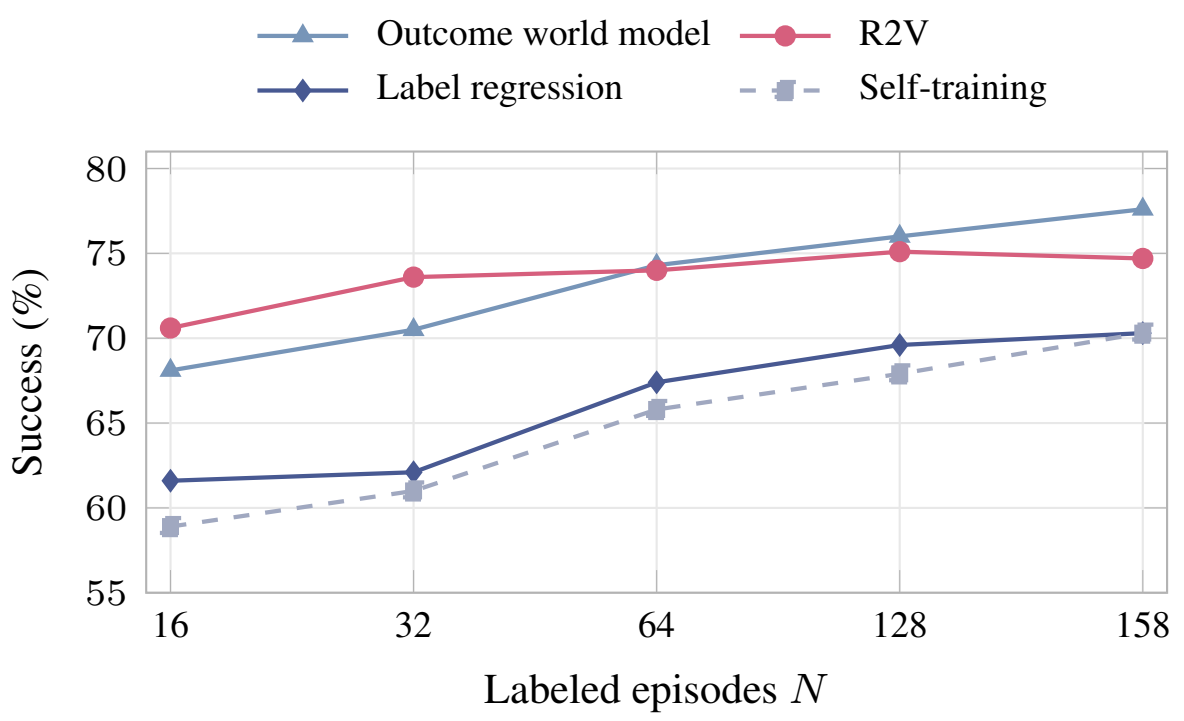

\paragraph{Label-budget comparisons.} \Cref{fig:budget} reports label-budget curves with the validation protocols in \cref{app:protocol}. At $N=158$, every execution is labeled and self-training coincides with label regression. R2V uses a graded reader target. At this full-label budget, replacing in-sample readings with out-of-fold scores retains 3.4 of the 4.4-point screening gain.

\paragraph{Label-subset replication.} The neural verifier is replicated across 20 label subsets, with intervals resampling subsets and episodes together. R2V gains 9.2 [5.3, 13.1] and 10.1 [6.2, 14.2] points at $N=16$ and 32 on screening episodes, and 9.6 [7.2, 12.1] and 9.3 [7.0, 11.7] on confirmation episodes. The encoder comparisons in \cref{app:encoders} use matched learners and labeled subsets within each configuration.

\subsection{Predicting the Recorded Feature}\label{app:owm}

\begin{table}[!htb]
\centering
\small
\setlength{\tabcolsep}{3pt}
\caption{\textbf{Outcome-feature prediction.} Success (\%) on 512 held-out episodes, scored as in \cref*{tab:main}, with 95\% episode-bootstrap intervals. Outcome-feature epochs use 32 external development scenes. Scalar R2V and label regression use within-training cross-validation. The four contrasts, registered after \cref{fig:budget}, use paired sign-flip tests and one Holm family. W:L counts episodes with positive and negative paired differences. These runs are separate from the recording-budget runs in \cref{tab:grid}.}
\label{tab:owm}
\begin{tabular}{@{}lcc@{}}
\toprule
Selector & $N=64$ & $N=158$ \\
\midrule
Label regression & 64.8 [60.7, 68.6] & 70.1 [66.7, 73.4] \\
\rowcolor{ours} R2V & \textbf{72.7} [68.9, 76.2] & 72.3 [68.9, 75.6] \\
Outcome world model & 72.0 [68.3, 75.7] & \textbf{75.1} [71.5, 78.6] \\
\midrule
\multicolumn{3}{@{}l}{\emph{Outcome world model minus (points)}} \\
Label regression & $+$7.2 [3.6, 10.8] & $+$4.9 [2.3, 7.6] \\
\quad Holm $p$, W:L & 0.0004, 78:37 & 0.0012, 113:59 \\
R2V & $-$0.7 [$-$3.4, 2.1] & $+$2.8 [0.5, 5.1] \\
\quad Holm $p$, W:L & 0.70, 40:40 & 0.036, 107:55 \\
\bottomrule
\end{tabular}
\end{table}

The feature-prediction variant applies the same outcome reader to its predicted representation. \Cref{tab:owm} reports registered confirmation comparisons at $N=64$ and 158.

\subsection{Separate Validation and Target Comparisons}\label{app:robust}
The external-validation comparison selects epochs on 32 labeled development scenes (seeds 160--191). The neural block in \cref{tab:trees} reports this target-comparison run after restricting selection to feasible candidates with recorded outcomes. At $N=16$, its direct, hard-target, and logit-target success rates are 59.7\%, 68.8\%, and 69.1\%. At $N=158$, they are 74.0\%, 73.4\%, and 75.3\%. All target variants in this block share the external-validation and candidate-selection protocol. The primary comparison uses within-training cross-validation.

\section{Recorded and Generated Futures}\label{app:push}

\subsection{PokeWorld Pushing}
Each of 48 PokeWorld~\cite{tan2026identifiability} episodes provides four candidate pushes. DreamDojo generates eight futures per candidate, and the simulator renders the corresponding executions. A distance-based readout measures final object-to-goal distance in both sources. It ranks all candidates correctly on simulator-rendered futures. On generated futures it selects the best push in 27.1\% of episodes [14.6, 39.6], against 25\% chance (\cref{tab:push}). The Spearman correlation on generated futures is 0.21 [0.08, 0.35].

\begin{table}[!htb]
\centering
\small
\setlength{\tabcolsep}{4pt}
\caption{\textbf{The same readout on two sources of future evidence.} PokeWorld, 48 episodes and four candidates. Top-1 is the fraction selecting the lowest-cost push. Top-pair is one minus the strict best--runner-up misordering rate on the 47 episodes with a unique true best. One episode with tied true costs is excluded (tolerance $10^{-8}$). Predicted ties are not counted as errors. Generated scores average eight DreamDojo samples per candidate.}
\label{tab:push}
\begin{tabular}{@{}lccc@{}}
\toprule
Future source & Spearman & Top-1 (\%) & Top-pair (\%) \\
\midrule
Simulator-rendered & 1.00 & 100.0 & 100.0 \\
DreamDojo-generated & 0.21 & 27.1 & 51.1 \\
\bottomrule
\end{tabular}
\end{table}

The position extractor is checked on the initial frame of all 1,536 generated videos. Its error relative to the simulator's object position is at most 0.8 pixels and 0.4 pixels on average. This validates the measurement at the observed state, while the matched future comparison tests whether predicted evolution preserves action ranking.

\subsection{World Action Models}\label{app:cosmos}
The Cosmos Policy diagnostic uses the released RoboCasa checkpoint, with five action-denoising steps, taking the first sampled action chunk at each decision. The task-selection criterion was within-scene outcome variation across sampling seeds. PnPCounterToSink, TurnOffStove, PnPCounterToCab, and OpenSingleDoor each use 30 scenes and four sampling seeds.

Readers use linear probes on frozen V-JEPA~2.1 features with eight-frame encoder inputs. Probes are fitted separately for imagined and recorded evidence using the same scene splits. For Cosmos, \cref*{tab:wam} uses scene-level leave-one-out scores, which exclude all sampled rollouts from the evaluated scene when fitting the probe. Value and imagined-future scores are taken at the first gripper-closing decision, while recorded scores use terminal observations. Across the four Cosmos tasks, this decision is available for 120, 113, 117, and 120 sampled traces, respectively. Within-scene AUCs average over the 10, 7, 5, and 5 scenes whose sampled outcomes differ. Fast-WAM uses 40 scenes (320 candidate rows), with 10 such scenes. Each of five label subsets contains 16 scenes, and AUC is averaged over held-out scenes with differing outcomes, then over the five subsets. Imagined evidence uses joint video-action inference (\texttt{infer\_joint}) at the first-contact decision on open\_microwave, with eight sampling seeds.

\paragraph{Evidence sampling.} For Cosmos, the imagined input is the predicted head-camera frame produced at decision $g$ for the sampled action chunk, repeated eight times for video encoding. The RoboCasa checkpoint predicts a 32-step action chunk and its future state, while evaluation executes the first 16 actions before replanning, so this frame targets the state 32 environment steps ahead. Recorded evidence uses the last eight observations retained by subsampling the stored trajectory at a stride of three. For Fast-WAM, the encoder uses the last eight frames of the stored terminal or generated clip. Short inputs are padded by repeating the first frame. Thus \cref*{tab:wam} compares decision-time predictive signals with terminal outcome evidence. The PokeWorld experiment uses an identical cost readout on the corresponding completed push in both sources.

\section{Theory}\label{app:theory}

\paragraph{Setup.} Draw an executed pair $(h,a)$ from the training distribution, and let $X$, $V$, and $Y$ denote its pre-action input, recorded window, and outcome. We allow $Y\in[0,1]$, which includes the binary success labels of \cref*{sec:method} as a special case. Norms are $L^2$ under this distribution, $\|u\|^2=\mathbb E\,u^2$. Write $f(X)=\mathbb E[Y\mid X]$, $\eta(X,V)=\mathbb E[Y\mid X,V]$, and $g^\star(V)=\mathbb E[Y\mid V]$. For any square-integrable function $u(X,V)$, define $(Tu)(x)=\mathbb E[u(X,V)\mid X=x]$. A video-only reader is the special case $u(x,v)=g(v)$, and $T\eta=f$.

\paragraph{Decomposition.} Conditioning on the pre-action input and recording splits the outcome into three orthogonal parts:
\begin{equation}\label{eq:decomp}
Y-\mathbb EY=\underbrace{f-\mathbb EY}_{\text{before acting}}+\underbrace{\eta-f}_{\text{in the recording}}+\underbrace{Y-\eta}_{\text{never shown}}.
\end{equation}
The terms are martingale differences for the successive information sets generated by $X$, $(X,V)$, and $(X,V,Y)$. Thus
\begin{gather*}
S=\mathrm{Var}\,f,\qquad
\Delta_{\mathrm{rec}}=\|\eta-f\|^2=\mathbb E\,\mathrm{Var}(\eta\mid X),\\
R=\|Y-\eta\|^2,\qquad S+\Delta_{\mathrm{rec}}+R=\mathrm{Var}\,Y.
\end{gather*}
For binary $Y$, $\Delta_{\mathrm{rec}}=0$ holds exactly when $Y\perp V\mid X$, or equivalently $I(Y;V\mid X)=0$. The decomposition also applies within each episode when all expectations are conditioned on that episode. For $\mathrm{Var}\,Y>0$, the optimal predictor from $(X,V)$ explains the share $(S+\Delta_{\mathrm{rec}})/\mathrm{Var}\,Y$, and one using $V$ alone explains at most that. Its explained variance can come from $S$, which a pre-action predictor could also capture, or from $\Delta_{\mathrm{rec}}$, which the recording reveals.

$\Delta_{\mathrm{rec}}$ is the optimal squared-loss reduction from observing the recording. When $\Delta_{\mathrm{rec}}+R>0$, the fraction $\nu=\Delta_{\mathrm{rec}}/(\Delta_{\mathrm{rec}}+R)$ measures how much of the uncertainty remaining after observing $X$ the recording resolves. Knowing $P(X)$ alone does not identify $f$ when $P(Y\mid X)$ is unconstrained~\cite{scholkopf2012causal}. A finite learner can nevertheless benefit from recordings when $\Delta_{\mathrm{rec}}=0$.

\paragraph{Hindsight without new information.} Let $X$ be uniform on $\{0,1\}^d$, $V=\langle\theta,X\rangle\bmod 2$, and $Y=V\oplus b\oplus\xi$, where $\xi\sim\mathrm{Bernoulli}(\rho)$ is independent noise, $\rho<\tfrac12$, and $\theta\in\{0,1\}^d$ and $b\in\{0,1\}$ are unknown. Since $V$ is a function of $X$, $\eta=f$ and $\Delta_{\mathrm{rec}}=0$. Majority vote on $Y\oplus V$ recovers the reader's bit $b$ from $O(\log(1/\delta)/(1-2\rho)^2)$ labels, independently of $d$. Meanwhile, $O(d+\log(1/\delta))$ recordings $(X,V)$ recover $\theta$ by elimination over $\mathrm{GF}(2)$. Direct prediction from $(X,Y)$ requires $\Omega(d)$ labels in the worst case for nontrivial accuracy at fixed noise and confidence. The label requirement for the reader is independent of $d$, while learning the relation between $X$ and $V$ requires a number of recorded pairs that grows with $d$.

\paragraph{Hindsight advantage at a label budget.} Let $R_X$ and $R_V$ be squared-loss risks on an independent test draw using $X$ and $V$, respectively. If $\eta(X,V)=g^\star(V)$, predictors $\hat f_n$ and $\hat g_n$ fitted on the same $n$ labels satisfy
\begin{equation*}
\begin{aligned}
&\mathbb ER_X(\hat f_n)-\mathbb ER_V(\hat g_n)\\
&\quad=\Delta_{\mathrm{rec}}+\mathbb E\|\hat f_n-f\|^2-\mathbb E\|\hat g_n-g^\star\|^2.
\end{aligned}
\end{equation*}
The outer expectations average over fitted models. The expression separates the information revealed by the recording from the estimation difficulty of reading versus predicting it.

\paragraph{Target noise and the labeled-only control.} The ideal target $\eta$ has the same conditional mean $f$ as $Y$, but $\mathbb E\,\mathrm{Var}(\eta\mid X)=\Delta_{\mathrm{rec}}$ whereas $\mathbb E\,\mathrm{Var}(Y\mid X)=\Delta_{\mathrm{rec}}+R$. It can therefore reduce target noise even without additional recordings. A fitted reader also introduces estimation error, so this population calculation does not guarantee a finite-sample gain. Our labeled-only control showed no improvement (\cref*{sec:labels}). The observed gain came from using reader targets on additional executions.

\paragraph{Where reader errors go.} Condition on a reader $g$ trained independently of the verifier sample and write $e=g-\eta$. Its population squared-loss regression target is $\mathbb E[g\mid X]=f+\mathbb E[e\mid X]$. Because
\begin{equation*}
\|e\|^2=\|\mathbb E[e\mid X]\|^2+\mathbb E\,\mathrm{Var}(e\mid X),
\end{equation*}
only the component predictable from $X$ remains in the population target. The remainder averages out. A video-only reader also carries the approximation error $g^\star(V)-\eta(X,V)$, which vanishes when $Y\perp X\mid V$. The main experiment uses in-sample scores for labeled executions, and an out-of-fold target comparison is reported in \cref{app:budget}.

\paragraph{Selection.} With $\hat a$, $a^\star$, and $H_K$ as in \cref*{sec:method}, adding and subtracting $\mathbb E\max_a f(a)$ gives \cref*{eq:selection}. Here $f$ and expectations refer to the evaluation distribution of scenes and candidate pools. Selection is invariant to positive affine score transformations. For any $\beta>0$ and $c\in\mathbb R$, set $e_h(a)=\beta\hat s(x(h,a))-c-f(x(h,a))$. Within each episode,
\[
f(x(h,a^\star))-f(x(h,\hat a))\le 2\max_a|e_h(a)|.
\]
Bounding the maximum by the root sum of squares, then applying Jensen's inequality, gives
\begin{align*}
\mathcal R_K
&\le 2\mathbb E_h\left[\sum_{a\in\mathcal A(h)}e_h(a)^2\right]^{1/2}\\
&\le 2\left[\mathbb E_h\sum_{a\in\mathcal A(h)}e_h(a)^2\right]^{1/2}.
\end{align*}
For fixed $K$, define $\|e\|_{\mathrm{cand}}^2=\mathbb E_h\frac1K\sum_{a\in\mathcal A(h)}e_h(a)^2$. Then $\mathcal R_K\le 2\sqrt K\,\|e\|_{\mathrm{cand}}$. This norm uses evaluation candidates rather than training executions. The preceding sum form also accommodates varying candidate counts.

\paragraph{World models and R2V.} Using $T$ and $\eta$ under the same evaluation distribution, a video world model estimates the conditional law of $V$ given $X$, inducing an approximation $\hat T$ to $T$. For a fixed ideal outcome estimate $\eta$, its induced score error is $\|(\hat T-T)\eta\|_{\mathrm{cand}}$. Averaging independent imagined samples reduces sampling variance around $\hat T\eta$. When the window is sufficient and the probability reader is ideal, $g^\star(V)=\eta(X,V)$, so R2V directly estimates $T\eta=f$, a value-equivalent target for selection~\cite{grimm2020value}.

\section{Additional Manipulation Tasks}\label{app:tasks}

\paragraph{Tasks and candidates.}
Place\_can\_basket, handover\_block, and stamp\_seal use RoboTwin. Adjust\_bottle uses DOMINO. Each scene offers eight scripted alternatives. RoboTwin candidates specify contact points, and cuRobo plans a path to each. DOMINO candidates specify interception times. Candidates with failed inverse kinematics or planning are infeasible. The fixed-order baseline ranks candidate indices by success in the labeled scenes and selects the first feasible candidate. The scripted expert selects its own contact or interception time.

The pre-action input has 881 coordinates: a 768-dimensional V-JEPA feature of the first head-camera frame, an eight-way candidate index, the planned outcome time, a contact index or interception time, 96 coordinates from the pre-action joint path resampled to 16 points, and a seven-coordinate end-effector pose. The reader uses the 1.2-second window starting at the planned outcome phase.

\paragraph{Data and evaluation.}
Each task has 150 training scenes with one execution per candidate and 100 screening scenes with three executions per candidate. RoboTwin uses seeds 10001--10150 for training, 10151--10250 for screening, and 10251--10650 for confirmation. DOMINO uses seeds 4001--4150, 4151--4250, and 4251--5250. Four planned confirmation scenes lack stored trajectories (4539, 4550, 4742, 5102), leaving 996. Neural results average 20 labeled subsets. Differences use intervals resampling scenes and label subsets. The eight neural confirmation contrasts at $N=16,32$ form one Holm family.

\begin{table*}[!t]
\centering\footnotesize
\setlength{\tabcolsep}{3pt}
\caption{\textbf{Four additional manipulation tasks.} Neural confirmation success (\%) at $N=16$, and the final column gives the separate $N=32$ contrast. Each chosen candidate is scored over three executions. Fixed ranks candidate indices using the labeled scenes. OWM predicts the outcome feature. Observed best selects and scores on the same three-run means. Headroom is the gain of direct regression from 16 to all 150 labeled training scenes. The last two columns report R2V minus direct regression with 95\% intervals.}
\label{tab:tasks}
\begin{tabular}{@{}lr rrrrrr r cc@{}}
\toprule
Task & Scenes & Expert & Fixed & LR & \cellcolor{ours}R2V & OWM & Obs. best & Headroom & $\Delta$, $N=16$ & $\Delta$, $N=32$ \\
\midrule
Place can basket & 400 & 74.0 & \textbf{74.1} & 69.8 & \cellcolor{ours}73.8 & 73.1 & 90.8 & 4.7 & $+4.1$ [1.9, 6.4] & $+4.1$ [2.5, 5.9] \\
Handover block & 400 & 89.3 & 95.3 & 96.4 & \cellcolor{ours}\textbf{96.8} & 96.7 & 97.9 & 0.6 & $+0.4$ [0.0, 0.9] & $-0.1$ [$-$0.6, 0.3] \\
Stamp seal & 400 & 61.5 & \textbf{66.7} & 64.3 & \cellcolor{ours}65.6 & 65.6 & 67.5 & 1.3 & $+1.3$ [0.5, 2.3] & $+0.3$ [$-$0.3, 0.9] \\
Adjust bottle & 996 & 45.1 & 61.8 & 58.1 & \cellcolor{ours}\textbf{63.0} & 62.6 & 77.9 & 6.6 & $+4.9$ [2.8, 7.3] & $+3.3$ [1.9, 4.8] \\
\bottomrule
\end{tabular}
\end{table*}

\paragraph{Learner and label budget.}
At $N=16$, tree R2V gains 3.1 points on place\_can\_basket, 10.8 on adjust\_bottle, $-0.3$ on handover\_block, and 1.6 on stamp\_seal. The corresponding tree label-limited headrooms are 3.5, 11.4, 0.1, and 2.4 points. In the 100-scene screening study, neural R2V minus direct regression with all 150 scenes labeled ranges from $-2.7$ to 0.0 across the four tasks. Confirmation results are reported separately in \cref{tab:tasks}. The fixed-order baseline is within 1.6 points of neural R2V in every task.

\paragraph{Training-data checks.}
The pilot evaluated seven tasks on 64 scenes each. Handover\_mic was excluded because none of its candidates succeeded. Move\_pillbottle\_pad and RoboTwin adjust\_bottle had differing candidate outcomes in only 14 and 3 of 64 scenes, below the 30\% disagreement criterion. The included DOMINO adjust\_bottle is a different task implementation.

The main study computed two checks on the 150 training scenes. A gain was predicted when the reader's within-scene margin over the pre-action predictor was at least 0.10 and direct regression gained at least three points from 16 to 150 labels. Each task received a prediction before any evaluation scene was scored. A confirmation gain required at least three points and Holm-adjusted $p<0.05$. Three predictions held (\cref{tab:checks}).

\begin{table}[!tb]
\centering\footnotesize
\setlength{\tabcolsep}{3pt}
\caption{\textbf{Training-data checks and confirmation outcomes.} Margin compares the reader and predictor. Headroom here uses one execution per training candidate. Predictions were fixed before evaluation.}
\label{tab:checks}
\begin{tabular}{@{}lrrll@{}}
\toprule
Task & Margin & Headr. & Prediction & Outcome \\
\midrule
Place can basket & 0.10 & 2.0 & No gain & Gain \\
Handover block & 0.00 & 0.0 & No gain & No gain \\
Stamp seal & 0.00 & 0.3 & No gain & No gain \\
Adjust bottle & 0.42 & 3.6 & Gain & Gain \\
\bottomrule
\end{tabular}
\end{table}

The separate headroom analysis in \cref*{fig:mechanism}c was registered after screening and before confirmation scoring. Place\_can\_basket had the largest execution disagreement, with 34\% of feasible candidates changing outcome across repeats, compared with 2\%, 14\%, and 13\% for handover\_block, stamp\_seal, and adjust\_bottle. Its single-run training headroom was 2.0 points, while three-run screening and confirmation headrooms were 6.2 and 4.7. Repeated outcomes therefore change the estimated room for learning in this task.

\section{Other Encoders}\label{app:encoders}

\begin{table}[!htb]
\centering
\small
\setlength{\tabcolsep}{4pt}
\caption{\textbf{Three encoders.} Success (\%) on the 128 screening episodes when one frozen encoder supplies both the pre-action scene feature and the reader's window feature. Each entry averages five label subsets. All rows use the same re-rendered clips and are compared within this encoder study, separately from the primary feature pipeline. Differences are computed before rounding.}
\label{tab:general}
\begin{tabular}{@{}lccc@{}}
\toprule
 & \multicolumn{3}{c}{Labeled episodes $N$} \\
\cmidrule(l){2-4}
Encoder and selector & 16 & 32 & 64 \\
\midrule
\multicolumn{4}{@{}l}{\emph{V-JEPA 2.1}~\cite{murlabadia2026vjepa21}} \\
\quad Label regression & 61.2 & 61.5 & 66.5 \\
\rowcolor{ours} \quad R2V & \textbf{67.6} & \textbf{72.0} & \textbf{71.7} \\
\quad Gain & $+$6.4 & $+$10.5 & $+$5.2 \\
\midrule
\multicolumn{4}{@{}l}{\emph{DINOv2 ViT-B/14}~\cite{oquab2024dinov2}} \\
\quad Label regression & 59.4 & 62.0 & 65.9 \\
\rowcolor{ours} \quad R2V & \textbf{69.7} & \textbf{74.1} & \textbf{71.6} \\
\quad Gain & $+$10.3 & $+$12.1 & $+$5.7 \\
\midrule
\multicolumn{4}{@{}l}{\emph{SigLIP}~\cite{zhai2023siglip} } \\
\quad Label regression & 59.2 & 62.0 & 66.1 \\
\rowcolor{ours} \quad R2V & \textbf{67.7} & \textbf{72.3} & \textbf{72.0} \\
\quad Gain & $+$8.5 & $+$10.3 & $+$5.9 \\
\bottomrule
\end{tabular}
\end{table}

\paragraph{Encoder comparison.} We replace V-JEPA with DINOv2 ViT-B/14 or SigLIP in both the reader and the pre-action input, keeping the action representation and verifier protocol fixed. At $N=16$, R2V improves over matched label regression by 6.4, 10.3, and 8.5 points, respectively (\cref{tab:general}). The corresponding gains at $N=32$ are 10.5, 12.1, and 10.3 points. The complete table includes $N=64$.

\section{Recording Budgets}\label{app:record}

\begin{table}[!tb]
\centering
\small
\setlength{\tabcolsep}{4.5pt}
\caption{\textbf{Label and recording budgets.} Single-choice success (\%) on the 512 held-out episodes with $n_L$ labeled training episodes and requested budgets $n_U$ of new scenes with one random execution each. Labels only: label regression on the $n_L$ labeled episodes. These recording-budget runs are separate from \cref{tab:owm}, so each table keeps its own direct baseline.}
\label{tab:grid}
\begin{tabular}{@{}rccccc@{}}
\toprule
 & \multicolumn{4}{c}{\cellcolor{ours}R2V, recordings $n_U$} & Labels \\
\cmidrule(lr){2-5}
$n_L$ & 142 & 270 & 540 & 1080 & only \\
\midrule
\multicolumn{6}{@{}l}{\emph{Gradient-boosted trees}} \\
16 & 73.8 & 73.1 & 73.9 & \textbf{74.8} & 69.9 \\
32 & 76.0 & 76.9 & \textbf{77.4} & 77.0 & 73.5 \\
64 & 78.7 & \textbf{79.2} & 78.8 & 79.0 & 77.6 \\
128 & 80.0 & 80.0 & 79.9 & \textbf{80.7} & 79.8 \\
\midrule
\multicolumn{6}{@{}l}{\emph{Neural verifier}} \\
16 & 63.4 & 63.6 & 66.1 & \textbf{68.2} & 59.2 \\
32 & 64.6 & 65.8 & 70.4 & \textbf{71.9} & 60.1 \\
64 & 69.8 & 71.5 & 73.6 & \textbf{76.2} & 65.0 \\
128 & 73.2 & 74.9 & 76.7 & \textbf{78.3} & 68.4 \\
\bottomrule
\end{tabular}
\end{table}

\begin{table}[!htb]
\centering
\footnotesize
\setlength{\tabcolsep}{2pt}
\caption{\textbf{Recording budget.} R2V gains (points) over label regression. Single-execution rows average five label subsets. Scenes gives source scenes at $N=16$. Single-execution rows name the requested budgets. Eval denotes screening (128) or confirmation (512). Bottom: registered comparisons against label regression (L) and self-training (S), in two four-contrast Holm families, all adjusted $p<0.001$. Confirmation seeds are 640--1151. Added training scenes are 2000--2141 (random) and 2000--2539 (logged).}
\label{tab:record}
\begin{tabular}{@{}lcccc@{}}
\toprule
Recorded & Scenes & Eval & $N{=}16$ & $N{=}32$ \\
\midrule
All feasible & 142 & 512 & $+$10.2 & $+$11.5 \\
All feasible & 142 & 128 & \plus9.0 & $+$11.5 \\
\midrule
One, random & 142 & 128 & \plus3.3 & \plus5.6 \\
One, random & 270 & 128 & \plus4.2 & \plus6.9 \\
One, random & 540 & 128 & \plus6.2 & $+$11.0 \\
One, random & 1080 & 128 & \plus8.0 & $+$10.9 \\
One, logged & 142 & 128 & \plus0.6 & \plus2.8 \\
One, logged & 270 & 128 & \plus1.0 & \plus5.8 \\
One, logged & 540 & 128 & \plus3.5 & \plus7.2 \\
One, logged & 1080 & 128 & \plus4.5 & \plus7.9 \\
\midrule
One, random, L & 142 & 512 & $+$4.2 [2.5, 6.0] & $+$4.5 [2.9, 6.0] \\
One, random, S & 142 & 512 & $+$4.1 [2.3, 5.9] & $+$5.3 [3.8, 6.9] \\
One, logged, L & 540 & 512 & $+$5.5 [3.5, 7.5] & $+$7.1 [5.2, 9.0] \\
One, logged, S & 540 & 512 & $+$6.1 [4.1, 8.1] & $+$8.2 [6.3, 10.1] \\
\bottomrule
\end{tabular}
\end{table}

\paragraph{Collection protocols.} Executing all feasible candidates supplies 2,503 unlabeled recordings at $N=16$ and 2,251 at $N=32$. The single-execution protocols instead add one recording per new scene, using either a random candidate or the analytic synchronization selector from \cref{app:baselines}. The logger does not depend on $N$. Among feasible candidates with a valid arrival time, it minimizes the lexicographic key $(|t_{\mathrm{arr}}-t_{\mathrm{int}}|,t_{\mathrm{int}},\text{index})$. \Cref{tab:record} gives the complete random and logged budget curves and their confirmation comparisons. At $N=16$, 142 random recordings yield a 4.2-point confirmation gain [2.5, 6.0], and 540 logged recordings yield 5.5 points [3.5, 7.5].

\paragraph{Two budgets.} The complete two-budget grid is reported in \cref{tab:grid}.

\section{Labels and Executions}\label{app:executions}

\Cref{tab:executions} counts every labeled and unlabeled execution, with labeled executions subsampled inside each source scene. Both single-execution analyses start from the requested 1,080-scene prefix. The execution accounting in \cref{tab:executions} retains the 1,071 scenes with a valid selected random candidate and rendered execution, while the candidate-budget analysis in \cref{app:recordingk} uses the 1,076 scenes with stored candidate draws. The two counts therefore correspond to their respective analysis filters.

\begin{table}[!htb]
\centering
\small
\setlength{\tabcolsep}{2.6pt}
\caption{Screening success (\%) against the executions each selector consumes, 128 episodes, five label subsets. Labeled executions are subsampled inside each labeled scene, nested across sizes. Unlabeled executions are one random candidate in each of 1,071 new scenes, except the first two R2V rows, which execute every candidate of the unlabeled training scenes. Reader AUC is on the 32 development episodes, whose executions are not counted.}
\label{tab:executions}
\begin{tabular}{@{}lrrrrcc@{}}
\toprule
Selector & $N$ & Lab. & Unlab. & Total & AUC & Succ. \\
\midrule
Label regression & 16 & 294 & 0 & 294 & -- & 61.6 \\
Label regression & 32 & 546 & 0 & 546 & -- & 62.1 \\
Label regression & 64 & 1{,}092 & 0 & 1{,}092 & -- & 67.4 \\
Label regression & 158 & 2{,}797 & 0 & 2{,}797 & -- & 70.3 \\
\midrule
\rowcolor{ours} R2V & 16 & 294 & 2{,}503 & 2{,}797 & 0.848 & 70.6 \\
\rowcolor{ours} R2V & 32 & 546 & 2{,}251 & 2{,}797 & 0.851 & \textbf{73.6} \\
\midrule
R2V & 16 & 16 & 1{,}071 & 1{,}087 & 0.641 & 54.5 \\
R2V & 16 & 58 & 1{,}071 & 1{,}129 & 0.791 & 64.2 \\
R2V & 32 & 32 & 1{,}071 & 1{,}103 & 0.742 & 61.5 \\
R2V & 32 & 63 & 1{,}071 & 1{,}134 & 0.794 & 65.5 \\
R2V & 32 & 122 & 1{,}071 & 1{,}193 & 0.834 & 70.6 \\
\bottomrule
\end{tabular}
\end{table}

\section{Direct Baselines}\label{app:baselines}

\begin{table}[!htb]
\centering
\small
\caption{\textbf{Direct-supervision baselines.} Success (\%) on 128 screening episodes, averaged over five label subsets. The learned methods share the neural verifier architecture. The analytic rule uses task geometry without learning. The synchronization variants add its score to the verifier input. On the 512 confirmation scenes, the rule reaches 71.0\% and neural R2V + synchronization reaches 68.9\% and 71.9\% at $N=16,32$. The registered contrasts against the rule have Holm-adjusted $p=0.31,0.50$.}
\label{tab:baselines}
\begin{tabular}{@{}lcc@{}}
\toprule
Selector & $N{=}16$ & $N{=}32$ \\
\midrule
Label regression, squared loss & 61.6 & 62.1 \\
Label regression, cross-entropy & 60.4 & 60.1 \\
Label regression, pairwise ranking & 59.9 & 59.2 \\
Label regression + synchronization & 64.6 & 63.3 \\
Analytic synchronization alone & \multicolumn{2}{c}{68.8} \\
\midrule
\rowcolor{ours} R2V & 70.6 & 73.6 \\
\rowcolor{ours} R2V + synchronization & \textbf{70.7} & \textbf{73.9} \\
\bottomrule
\end{tabular}
\end{table}

Squared loss is the strongest direct-supervision loss in \cref{tab:baselines} and is used in the main experiments. The synchronization rule selects, among feasible candidates with a valid arrival time, the lexicographic minimum of $(|t_{\mathrm{arr}}-t_{\mathrm{int}}|,t_{\mathrm{int}},\text{index})$, where $t_{\mathrm{arr}}$ is planned gripper arrival and $t_{\mathrm{int}}$ is bottle intercept time. Its score is an additional input in the synchronization variants.

\needspace{6\baselineskip}
\section{Learners and Targets}\label{app:trees}

\Cref{tab:trees} compares reader targets under a matched learner and training protocol within each block, including the full-label endpoint.

\begin{table}[!htb]
\centering
\small
\setlength{\tabcolsep}{3pt}
\caption{\textbf{Learners and supervision targets.} Success (\%) on the 512 confirmation episodes. Hard thresholds the reader probability at 0.5. Neural rows use external validation, while trees use fixed settings. Each learner block compares targets under one fixed protocol.}
\label{tab:trees}
\begin{tabular}{@{}llccc@{}}
\toprule
Learner & Target & $N=16$ & $N=32$ & $N=158$ \\
\midrule
Neural & label & 59.7 & 60.2 & 74.0 \\
 & reading, hard & 68.8 & 70.1 & 73.4 \\
\rowcolor{ours} & reading, logit & \textbf{69.1} & \textbf{71.2} & \textbf{75.3} \\
Trees & label & 69.9 & 73.5 & \textbf{81.2} \\
 & reading, hard & 74.6 & 76.6 & 80.5 \\
 & reading, probability & 75.4 & 77.1 & 80.6 \\
\rowcolor{ours} & reading, logit & \textbf{75.5} & \textbf{77.6} & 79.9 \\
\bottomrule
\end{tabular}
\end{table}

The neural block uses the external-validation protocol in \cref{app:robust} and compares the label, hard-reader, and logit-reader targets used in that study. The tree block additionally evaluates probability targets under its fixed learner settings.

\needspace{6\baselineskip}
\section{A Pretrained Vision-Language Reader}\label{app:vlm}

Cosmos-Reason2 (8B)~\cite{nvidia2025cosmosreason2} is prompted with: ``Look at the bottle and the gripper. Is the bottle held in the gripper? Answer yes or no.'' Each stored eight-frame window is supplied as one video input with frame resampling disabled. Selection uses 32 labeled development episodes and compares 30 configurations (five models, two windows, and three prompts). The reported setting uses the late window and the gripper prompt. The student target is the raw yes-token logit.

The VLM scorer operates on the stored late-window subset, yielding 2,778 scored clips from the 158 training episodes. Verifier fitting uses these VLM scores without success labels. Four-fold cross-validation over the scored training episodes selects 33 epochs, after which an ensemble with seeds 0, 1, and 2 is fit on all scored rows. The neural verifier reaches 68.6\% [64.7, 72.3] on all 512 confirmation episodes, a paired $+9.4$ points [5.9, 12.7] over the main $N=16$ label-regression baseline. The tree verifier reaches 76.4\% [72.8, 79.8] using scikit-learn gradient-boosted regression trees with 100 estimators, maximum depth 3, learning rate 0.05, and random seed 0.

\begin{table*}[!t]
\centering\small
\setlength{\tabcolsep}{3.5pt}
\caption{\textbf{Complete closed-loop PushT results.} All arms share 100 initial conditions (seeds 598000--598099). $K=1$ is the native policy, and $K>1$ uses R2V. Differences and intervals are in pp versus native, and W:L:T counts episode wins, losses, and ties. Pointwise percentile intervals use 5,000 paired-episode resamples with NumPy random seed 0. Reward is mean maximum environment reward, Overrides counts decisions per episode that select a candidate other than zero, and Seconds is mean episode wall time. Exact $p$ uses two-sided McNemar tests, and Holm adjusts the four comparisons with native.}
\label{tab:pusht_full}
\begin{tabular}{@{}rcrlcrrrrr@{}}
\toprule
$K$ & Success & $\Delta$ (pp) & 95\% CI (pp) & W:L:T & Reward & Overrides & Seconds & Exact $p$ & Holm $p$ \\
\midrule
1 & 42/100 & -- & -- & -- & 0.801 & 0.00 & 23.5  & -- & -- \\
2 & 41/100 & $-1.0$ & [$-10$, 7] & 9:10:81 & 0.800 & 15.27 & 45.2  & 1.0000 & 1.0000 \\
4 & 42/100 & $0.0$ & [$-9$, 9] & 11:11:78 & 0.789 & 22.72 & 91.8  & 1.0000 & 1.0000 \\
8 & 40/100 & $-2.0$ & [$-11$, 7] & 10:12:78 & 0.748 & 27.93 & 187.7  & 0.8318 & 1.0000 \\
16 & \textbf{51/100} & $+9.0$ & [1, 17] & 14:5:81 & 0.810 & 28.84 & 351.0  & 0.0636 & 0.2543 \\
\bottomrule
\end{tabular}
\end{table*}

\section{Closed-Loop PushT Evaluation}\label{app:closedloop}

\begin{figure}[!tb]
\centering
\begin{tikzpicture}
\begin{axis}[
 width=0.98\columnwidth,height=1.78in,
 tick label style={font=\scriptsize},label style={font=\small},
 legend cell align=left,grid=major,grid style={gray!18},
 axis line style={gray!60},tick style={gray!60},
 xlabel={Candidates $K$},ylabel={Success (\%)},
 ymin=35,ymax=76,ytick={40,50,60,70},
 xmin=0.8,xmax=5.2,xtick={1,2,3,4,5},xticklabels={1,2,4,8,16},
 legend style={at={(0.5,1.04)},anchor=south,legend columns=2,font=\scriptsize,draw=none,fill=none,column sep=4pt},
]
\addplot+[cRead,mark=*,line width=1pt,mark size=1.8pt] coordinates {(1,42.000000000) (2,41.000000000) (3,42.000000000) (4,40.000000000) (5,51.000000000)};
\addlegendentry{R2V}
\addplot[cGray,dashed,line width=0.8pt] coordinates {(1,42) (5,42)};
\addlegendentry{Native policy}
\node[font=\scriptsize,anchor=south west,text=cRead] at (axis cs:4.88,51) {51};
\end{axis}
\end{tikzpicture}
\caption{\textbf{Closed-loop PushT.} Repeated chunk selection with one frozen ranker on 100 shared initial conditions. The reader uses 64 labeled episodes. The dashed line is the native policy, and all candidate budgets are shown. Paired intervals and tests are in \cref{tab:pusht_full}.}
\label{fig:pusht}
\end{figure}
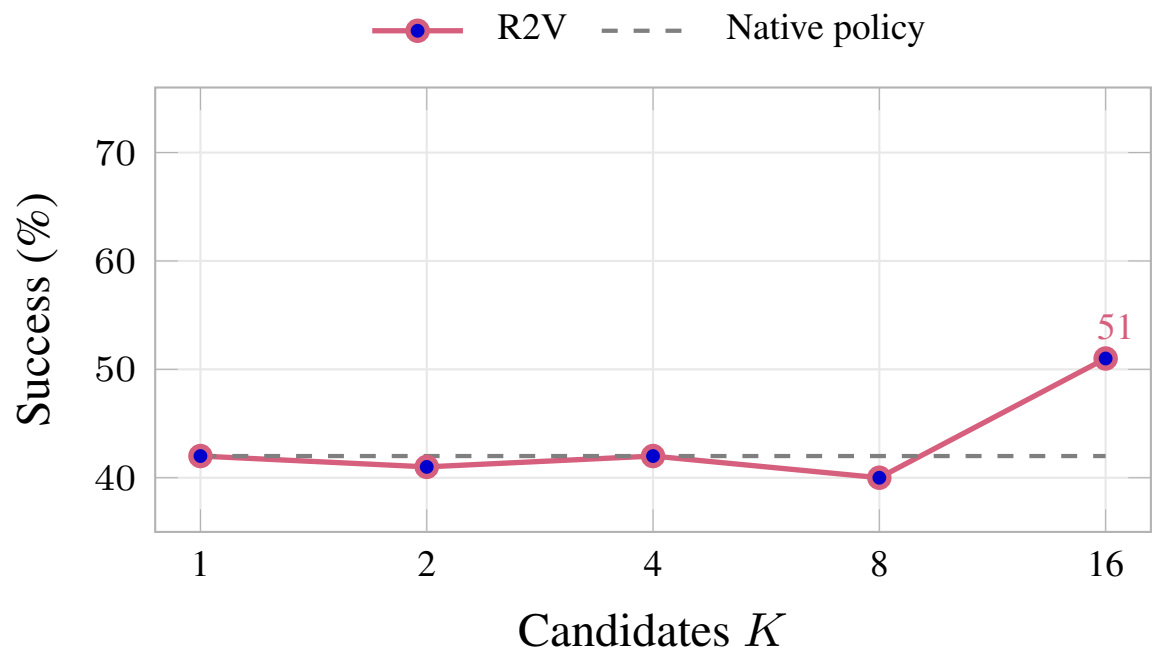

\paragraph{Overview.}
For closed-loop control, we replay a simulator prefix to the same state, execute each candidate chunk, and continue with the base policy. A frozen reader scores each branch's terminal observation. We train a state--action ranker on reader-score differences between candidates executed from the same state. At each test-time decision, the ranker selects the chunk with the largest ensemble-mean score, executes it, and repeats from the new observation. Success is 42\%, 41\%, 42\%, 40\%, and 51\% at $K=1,2,4,8,16$ (\cref{fig:pusht}), and mean episode runtime rises from 23.5 to 351 seconds (\cref{tab:pusht_full}).

\paragraph{Policy and task.}
We use the frozen CNN-image Diffusion Policy checkpoint for PushT~\cite{chi2023diffusion}. Observations are $96\times96$ RGB images and the two-dimensional agent position. Each executed chunk has eight absolute-position actions. The environment reports task success when object coverage exceeds 0.95. Mean maximum environment reward is reported separately from binary success. The $K=1$ arm is the native policy.

\paragraph{Reader.}
The frozen outcome reader uses V-JEPA~2.1 ViT-B features from a temporally subsampled terminal clip. For each clip, per-tubelet spatial-mean tokens are summarized by their mean, last token, last-minus-first difference, and standard deviation. A class-balanced logistic regression on standardized features produces the terminal outcome score. The reader is fit on the 64 labeled source episodes and then frozen for branch scoring.

\paragraph{Candidate-specific hindsight data.}
The reader is fitted on 64 labeled episodes from a 120-episode base-policy collection (seeds 520000--520119). The other 56 episodes provide three anchor states each. For each of four sampled candidates, we reset the branch environment with the source episode's seed and replay the deterministic candidate-zero prefix to its anchor. We then execute that candidate's eight-action chunk and continue with candidate zero from the base policy until termination. This produces 168 states and 672 completed candidate branches. The frozen reader scores the terminal recording of each branch. A candidate receives only the target from its own execution, and simulator success flags are not used as ranker-fitting targets. Collection uses reset and prefix replay, without restoring a live environment from a snapshot.

\paragraph{Ranker.}
The regularized pairwise bilinear ensemble uses the frozen policy encoder's 132-dimensional scene feature, the two-dimensional agent position, and the executed $8\times2$ action chunk relative to that position. Scene features are reduced to 16 PCA dimensions and concatenated with position. Standardized action features and explicit state--action cross-features are then fitted with ridge regression ($\alpha=10$). The target is the within-state difference in frozen reader logits, scaled by one training-pool standard deviation. Three members bootstrap whole source episodes and their scores are averaged. A fixed 40/16 source-episode split selects the bilinear architecture, followed by a refit on all 56 episodes. The final selector executes the candidate with the highest mean ensemble score, with the same frozen model and rule for every evaluation budget.

\paragraph{Evaluation protocol.}
The 100 final seeds (598000--598099) are disjoint from the training collection (520000--520119) and the development cohorts (593000--593023 and 594000--594049). Candidate noise is keyed by environment seed, decision index, and candidate index, without $K$, so increasing the budget preserves the candidate prefix at a common input state. Candidate-zero and prefix consistency are checked before analysis, and every arm has one complete rollout per final seed. The 24-seed cohort 593000--593023 was used to compare execution rules, and the final evaluation uses the frozen highest-score rule. The 50-seed cohort 594000--594049 provides additional development checks. Online selection scores candidate actions directly and neither simulates candidate futures nor restores the live environment.

\begin{table}[!tb]
\centering\small
\setlength{\tabcolsep}{3pt}
\caption{\textbf{Within-ranker comparisons on PushT.} First minus second budget on the same 100 final episodes. Differences and pointwise paired bootstrap intervals are in pp. Exact two-sided McNemar $p$ values are unadjusted.}
\label{tab:pusht_budget_pairs}
\begin{tabular}{@{}lrrlr@{}}
\toprule
Comparison & $\Delta$ (pp) & W:L:T & 95\% CI (pp) & Exact $p$ \\
\midrule
$K16-K4$ & $+9.0$ & 17:8:75 & [$-1$, 19]  & 0.1078 \\
$K16-K8$ & $+11.0$ & 17:6:77 & [2, 20]  & 0.0347 \\
$K2-K4$ & $-1.0$ & 11:12:77 & [$-10$, 8]  & 1.0000 \\
$K4-K8$ & $+2.0$ & 13:11:76 & [$-8$, 12]  & 0.8388 \\
$K2-K16$ & $-10.0$ & 6:16:78 & [$-19$, $-1$]  & 0.0525 \\
\bottomrule
\end{tabular}
\end{table}

\paragraph{Statistical tests.}
For completeness, \cref{tab:pusht_full} reports two-sided exact McNemar tests and Holm adjustment~\cite{holm1979} across the four comparisons with the native policy, alongside the paired bootstrap intervals. At $K=16$, exact $p=0.0636$ and Holm-adjusted $p=0.2543$. The bootstrap interval and exact paired test summarize the same 100 paired episodes with different statistics. Within-ranker exact tests in \cref{tab:pusht_budget_pairs} are unadjusted.

\paragraph{Interpretation.}
In closed-loop PushT, verifier-based reranking improved success at the largest tested candidate budget, from 42\% for the native policy to 51\% at $K=16$. Smaller candidate budgets did not improve success, so we interpret the result as evidence that learned candidate judgment can translate into a closed-loop gain rather than as monotonic closed-loop scaling. This experiment evaluates the closed-loop selector end to end, while the grasping experiments provide the matched supervision-source comparisons. Branch targets use base-policy continuation, whereas evaluation repeatedly selects with the verifier.

\Needspace{5\baselineskip}
\section{Recording Budget by Candidate Count}\label{app:recordingk}
\paragraph{Data and learners.}
Direct regression and R2V reuse stored grasp data with five shared 16-episode label subsets and GBDT settings (100 estimators, depth 3, learning rate 0.05). R2V adds one random execution per scene from nested prefixes starting at seed 2000. B142, B540, and B1080 name the requested prefixes, of which 141, 537, and 1,076 recordings are available. Missing seeds are 2002, 2192, 2532, and 2861 as they enter the prefix. This differs from the all-feasible pool of \cref*{fig:tts}.

\begin{table}[!tb]
\centering\footnotesize
\setlength{\tabcolsep}{2.5pt}
\caption{\textbf{Recording budget by candidate count.} GBDT grasp success (\%) on 195 scenes, using the same 16-episode label subsets. B142/B540/B1080 contain 141/537/1,076 recordings. Gain is $K16-K4$, in points.}
\label{tab:recording_k_full}
\begin{tabular}{@{}lrrrrrrr@{}}
\toprule
Model & $K1$ & $K2$ & $K4$ & $K8$ & $K16$ & All & Gain \\
\midrule
LR16 & 47.47 & 54.42 & 58.20 & 59.55 & 59.53 & 59.62 & 1.33 \\
B142 & 47.47 & 56.64 & 62.16 & 65.01 & 66.40 & 66.67 & 4.24 \\
B540 & 47.47 & 56.68 & 61.83 & 65.22 & 66.96 & 67.32 & 5.13 \\
\rowcolor{ours} B1080 & 47.47 & \textbf{57.71} & \textbf{63.72} & \textbf{67.33} & \textbf{68.91} & \textbf{69.23} & \textbf{5.19} \\
\bottomrule
\end{tabular}
\end{table}

\paragraph{Evaluation.}
Each of the same 195 confirmation scenes has at least 16 feasible candidates. We average the five label-subset results and 50 candidate-set draws (NumPy random seed 0) within each scene. All methods agree at $K=1$. Paired percentile intervals use 10,000 scene resamples, not draw-level resampling. The bootstrap uses one NumPy random generator initialized with seed 1, first for $D_{16}$ and then for the interaction.

\paragraph{Definition of the gain.}
The gain column in \cref{tab:recording_k_full} uses $K=16$ minus $K=4$ on the 195-scene recording-budget analysis. Main-paper \cref*{sec:labels} instead reports the full-candidate-pool minus $K=4$ contrast from the primary candidate-budget evaluation in \cref*{fig:tts}.

\paragraph{Paired contrasts.}
B1080 exceeds B142 at $K=16$ by $D_{16}=2.507$ points [0.443, 4.638], with scene-level W:L:T of 75:54:66. The difference in the $K=4$ to $K=16$ gain is $I=0.950$ points [$-0.704$, 2.585], with W:L:T of 79:83:33. More recordings improve $K=16$ success, although the interaction interval includes zero.

\end{document}